\documentclass[lettersize,journal]{IEEEtran}
\usepackage{amsmath,amsfonts}
\usepackage[ruled,vlined,linesnumbered]{algorithm2e}
\usepackage{array}
\usepackage[caption=false,font=footnotesize,labelfont=rm,textfont=rm]{subfig}
\usepackage{textcomp}
\usepackage{stfloats}
\usepackage{url}
\usepackage{verbatim}
\usepackage{graphicx}
\usepackage[percent]{overpic}
\usepackage{cite}
\usepackage{xspace}
\usepackage{makecell}
\usepackage{multirow}
\usepackage{multicol}
\usepackage{capt-of}
\usepackage{subscript}
\usepackage{pifont}
\usepackage{booktabs}
\usepackage{bbm}
\usepackage[colorlinks=true, linkcolor=blue, urlcolor=blue, citecolor=blue]{hyperref}

\usepackage{enumitem}

\newcounter{RNum}

\newcommand{\fref}[1]{Fig.~\ref{#1}}
\newcommand{\sref}[1]{Section~\ref{#1}}
\newcommand{\tref}[1]{Table~\ref{#1}}

\newcommand{\algref}[1]{Algorithm~\ref{#1}}

\newcommand{\shortname}{iFlax\xspace}

\usepackage[amssymb]{SIunits}
\usepackage{threeparttable}
\usepackage{array}
\newcommand{\PreserveBackslash}[1]{\let\temp=\\#1\let\\=\temp}
\newcolumntype{C}[1]{>{\PreserveBackslash\centering}p{#1}}
\newcolumntype{R}[1]{>{\PreserveBackslash\raggedleft}p{#1}}
\newcolumntype{L}[1]{>{\PreserveBackslash\raggedright}p{#1}}

\begin{document}

\title{Neuro-Symbolic Learning for Long-Horizon\\Task Planning Under Complex Logical Constraints}

\ifdefined\PaperAnonymous
  \author{Anonymous Submission (Source code will be released.)
          % <-this % stops a space
  }
\else
  \author{Qiwei Du$^{1}$, %427993
Zitong Zhan$^{1}$, %351065
Shaoshu Su$^{1}$, %310547
Bowen Li$^{2}$, %290727
Yi Du$^{1}$, %358477
Zhipeng Zhao$^{1}$, %346049
Taimeng Fu$^{1}$,\\ %311960
Sebastian Scherer$^{2}$, %104304
Jiaoyang Li$^{2}$, and %270001
Chen Wang$^{1}$ %200000

\thanks{Corresponding Email: \texttt{\{qiweid, chenw\}@sairlab.org}}%
\thanks{$^{1}$Spatial AI \& Robotics (SAIR) Lab, University at Buffalo, NY 14260}%
\thanks{$^{2}$Robotics Institute, Carnegie Mellon University,  PA 15213}%
% \thanks{$^{2}$AirLab, Carnegie Mellon University.}%
% \thanks{$^{3}$ARCS Lab, Carnegie Mellon University.}%
}
\fi

\ifdefined\PaperTROHeader
  \markboth{IEEE Transactions on Robotics%, Journal of \LaTeX\ Class Files,~Vol.~14, No.~8, August~2021
  }%
  {%Shell \MakeLowercase{\textit{et al.}}: A Sample Article Using IEEEtran.cls for IEEE Journals
  }
\fi


\makeatletter
\g@addto@macro\@maketitle{
  \begin{center}
    \setlength{\abovecaptionskip}{6pt}
    \setlength{\belowcaptionskip}{0pt}
    \includegraphics[width=1\linewidth]{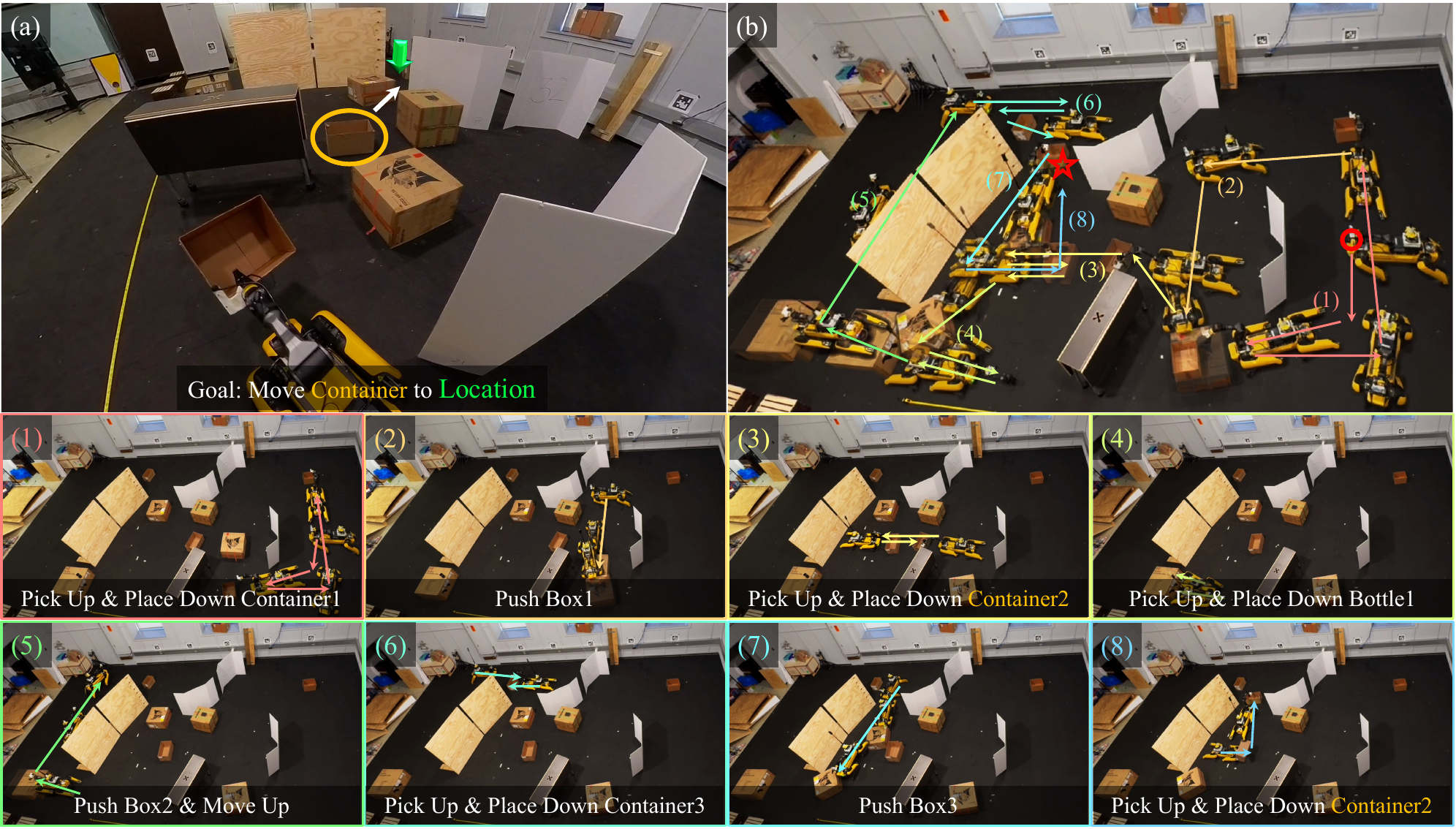}
    \setcounter{figure}{0}
    \vspace{-20pt} 
    \captionof{figure}{\textbf{Long-horizon task planning.} Given ``\textit{move container to location}'' in (a), the robot must satisfy complex logical constraints on fragile bottles, light containers, heavy boxes, and single-object manipulation. The execution trace in (b), with keyframes (1-8), shows a task requiring 51 ordered skills, each of which may involve multiple steps, e.g., a skill \textit{picking up a container} requires the robot to approach, sit down, detect, grasp, stand up, and retry after failure.}
    \label{fig:teaser}
  \end{center}
  \vspace{-25pt} 
}
\makeatother

\maketitle

\begin{abstract}
Task planning often suffers from severe efficiency bottlenecks when robots must reason over long-horizon action sequences under complex logical constraints, including object affordances, spatial relationships, and sequential action dependencies. Recent neuro-symbolic methods improve planning efficiency by learning object-importance scores to prune task-irrelevant objects, but they typically rely on fixed offline supervision generated from full search spaces. This creates a train-test mismatch: at deployment, the planner operates in pruned search spaces induced by the model's own imperfect predictions, leading to exposure bias and degraded planning performance. To address this challenge, we formulate object-importance learning for task planning as an imperative learning-based bilevel optimization problem. The upper level optimizes a neural scorer, while the lower level solves a symbolic planning problem in the score-pruned search space. To stabilize this learning process, we introduce a \textit{3R strategy} into the lower-level planning, using parallel \textit{Repair}, \textit{Restart}, and \textit{Rollback} recovery to provide reliable and adaptive feedback for upper-level learning. Experiments on three challenging benchmarks demonstrate state-of-the-art performance, including an 80.04\% reduction in failure rate and a 57.14\% reduction in planning time. We further validate the framework on a quadruped-based mobile manipulator in simulation and the real world, demonstrating its potential for efficient and deployable neuro-symbolic task planning.
\end{abstract}

\begin{IEEEkeywords}
Long-Horizon Task Planning, Neuro-Symbolic Learning, Mobile Manipulation, Imperative Learning.
\end{IEEEkeywords}
\vspace{-10pt}

\section{Introduction}
\IEEEPARstart{T}{ask} planning enables robots to achieve high-level goals by synthesizing a sequence of discrete actions that satisfy logical constraints.
In mobile manipulation, for example, delivering a beverage may require navigating to the kitchen, locating and grasping a bottle, transporting it to a desk, and rearranging obstacles that block execution. This capability is essential for various robotic applications such as household assistance~\cite{wang2020home,murray2022following}, logistics operations~\cite{li2024efficient,leet2023task,zhang2025real}, and search-and-rescue~\cite{yanmaz2023joint,osooli2024multi,romero2024cellular}. However, long-horizon task planning remains challenging because robots must reason over a large number of objects while accounting for \textit{complex logical constraints}, e.g., object affordances (bottles are pickable, but not pushable), spatial relationships (bottles may be placed on boxes, but not vice versa), and sequential action dependencies (bottle must be removed before pushing the box below it).
Here and throughout the paper, an ``object'' denotes a symbolic entity, which may be a non-physical entity such as a location, beyond manipulable physical items, per the definition of Planning Domain Definition Language (PDDL) \cite{mcdermott1998pddl,edelkamp2004pddl2}.
As the number of objects and logical constraints increases, the symbolic search space grows combinatorially \cite{fikes1971strips}, making planning efficiency a central bottleneck for real-world applications.

A common approach to such long-horizon complex task planning is to convert the robot's environment and task objective into a discrete optimization problem, where symbolic planners or graph-search algorithms find a feasible action sequence. However, the search space can be extremely large: an increasing number of objects expands the space of possible actions, while complex logical constraints impose long dependencies among these actions, leading to severe combinatorial explosion~\cite{garrett2021integrated,bylander1994computational}. Consequently, even state-of-the-art (SOTA) symbolic planners, such as LAMA planner~\cite{richter2010lama}, can become inefficient for long-horizon real-world execution.
For example, LAMA requires over 100 seconds to solve a challenging 15$\times$15 Maze Navigation among Movable Obstacles (MazeNamo) task~\cite{du2026fast}, where the robot must reason over both obstacle rearrangement and item-specific interaction constraints, such as which items can be picked, placed, or pushed.
These constraints substantially enlarge the feasible action space, making planning far more difficult than simple navigation in open space.
Figure~\ref{fig:teaser} illustrates this challenge in a real-world mobile manipulation task, where placing a container at its goal location requires long-horizon task planning over multiple interactions with various movable obstacles.

To mitigate this combinatorial growth, a natural strategy is to reduce the search space while preserving the objects necessary for a feasible solution. Recent neuro-symbolic methods (PLOI~\cite{silver2021planning} and Flax~\cite{du2026fast}) achieve this by pruning unimportant objects that are less likely to appear in the final plan. They learn \textit{object-importance scores} from planning data, rank objects by predicted importance, and progressively expand the planning problem using only the most important objects. This \textit{object-pruned planning} can substantially accelerate search; for example, in the same 15$\times$15 MazeNamo tasks requiring around 60 symbolic actions to complete, Flax~\cite{du2026fast} reduces planning time from over 100 seconds to about 24 seconds. However, imperfect importance scores may still cause the planner to \textit{miss critical objects} or \textit{include irrelevant ones}, leading to unsolvable or unnecessarily large subproblems. This remains a major bottleneck for long-horizon task planning.

We observe that a key reason for this performance degradation is the train-test mismatch. Existing methods~\cite{silver2021planning,du2026fast} typically train the importance scorer using fixed supervision generated offline by symbolic planners in the full search space. 
However, at test time, the planner operates in a pruned search space determined by the scorer's own predictions. 
This creates the notorious \textit{exposure bias}~\cite{ranzato2015sequence}: a model is trained under one distribution of inputs or states, but must operate at test time under states induced by its own predictions. 
However, in long-horizon planning, even small scoring errors can accumulate over expansion steps, leading to inefficient search or failure to find feasible solutions. Therefore, a more robust and efficient learning framework for long-horizon task planning is required to close this train-test gap, especially for the distribution of the task planning problems it will encounter at deployment.

To address this challenge, we propose to introduce imperative learning~\cite{wang2025imperative} and formulate object-importance learning for task planning as a bilevel optimization problem. The upper level optimizes a neural scorer to predict object importance, while the lower level solves the task-planning problem in the pruned search space induced by the scorer's current predictions. Unlike the prior SOTA method \cite{du2026fast} that learns from fixed offline supervision, we integrate the symbolic planner into the learning loop, allowing feasible plans generated by the planner to provide adaptive pseudo-supervision for the scorer. Intuitively, the neural scorer is exposed and trained under the same dynamically pruned search spaces it will encounter at test time, reducing exposure bias and better aligning object-importance learning with the actual planning objective.

Although online bilevel optimization is promising for addressing the \textit{exposure bias}, applying it to task planning remains challenging because the learning process can be fragile and unstable. In particular, during early training, the neural scorer may not yet provide meaningful object-importance predictions. As a result, the lower-level planner must search within an unreliable pruned space, where critical objects may be missing and misleading objects may be retained. This can prevent the planner from finding feasible plans and from returning meaningful adaptive feedback for upper-level score learning. To stabilize this process, we design a parallel \textit{3R strategy}, namely \textit{Repair}, \textit{Restart}, and \textit{Rollback} recovery strategies for the lower-level symbolic planning. This strategy enables the planner to provide reliable feedback for the upper-level learning by recovering from various early-stage failures.

We evaluate the proposed framework on three challenging benchmarks: \textit{MazeNamo}, \textit{SokoMindPlus}, and \textit{LogisticsPlus}. Across these benchmarks, our method achieves SOTA performance under large object sets, long horizons, and complex logical constraints. In particular, on the \textit{MazeNamo} benchmark, it reduces the failure rate by \textbf{80.04\%} and planning time by \textbf{57.14\%} compared with the prior SOTA method~\cite{du2026fast}. Beyond these benchmarks, we build a quadruped-based mobile manipulator and validate long-horizon \textit{MazeNamo} execution in both simulation and the real world. These results show that the proposed bilevel object-importance learning and 3R strategy not only improve planning efficiency, but also demonstrate the deployability of neuro-symbolic task planning on real robots.
For brevity, we refer to our method as \textbf{\shortname} to highlight our contribution in stabilizing the \textbf{i}mperative learning \cite{wang2025imperative} process for task planning and solving the \textit{exposure bias} of the prior SOTA method Flax \cite{du2026fast} under complex logical constraints.

In summary, this paper makes the following contributions:
\begin{itemize}
\item We propose \shortname, a neuro-symbolic task planning framework that formulates object-importance learning as a bilevel optimization process. By training the neural scorer through adaptive pseudo-supervision in score-pruned search spaces, \shortname overcomes the exposure bias of prior object-pruned planning methods.
\item We design a \textit{3R strategy} with parallel \textit{Repair}, \textit{Restart}, and \textit{Rollback} recovery in the lower-level symbolic planner. This strategy stabilizes bilevel optimization by helping the planner recover feasible solutions from unreliable early-stage pruned search spaces, thereby providing more reliable adaptive feedback for upper-level neural learning.
\item We demonstrate state-of-the-art performance on three challenging benchmarks and show substantial reductions in both failure rate and planning time. We further validate the deployability of \shortname on a quadruped-based mobile manipulator in both simulation and the real world.
\end{itemize}

\section{Related Work}
\subsection{Symbolic Task Planning in Large-Scale Environments}

To situate our contribution, we first review classical symbolic task planning in large-scale environments. This line of work exposes the combinatorial bottleneck most directly, which is precisely the difficulty our method aims to alleviate.

In this literature, long-horizon task planning is commonly modeled in PDDL \cite{mcdermott1998pddl,edelkamp2004pddl2}. Across these settings, the central bottleneck is combinatorial explosion: as the number of objects and possible interactions grows, the grounded search space expands rapidly \cite{bylander1994computational}. Classical work has mainly addressed this bottleneck by improving search heuristics, exploiting problem structure, or coupling symbolic planning with execution.

Within this setting, forward heuristic search remains a standard approach. Planners such as Fast Forward and Fast Downward use best-first search, enforced hill climbing, and domain-independent heuristics to prioritize promising states \cite{helmert2006fast,hoffmann2001ff,richter2010lama,helmert2009landmarks}. Early rule-based relevance analysis further filters obviously irrelevant objects before search begins \cite{nebel1997ignoring}. These methods often improve search order, but they do not fundamentally reduce the object set that defines the grounded planning problem. 
Beyond heuristic search, other paradigms reduce search through structural reasoning. Planning graphs prune inconsistent actions through mutex relations \cite{blum1997fast}, while satisfiability-based planners compile bounded-horizon planning into Boolean constraint-solving problems \cite{kautz1992planning,kautz2006satplan,kautz1999unifying}. These formulations have been effective in several classical domains, but they often struggle with the geometric and numerical constraints that arise in robotics \cite{garrett2021integrated,dantam2016incremental}. Hierarchical Task Network (HTN) planning reduces search depth by decomposing tasks into smaller subtasks \cite{nau2003shop2}, yet large environments with many objects can still make decompositions difficult to design and expensive to search \cite{georgievski2015htn}.

At the system level, prior robotic platforms also embed symbolic planners into execution frameworks \cite{cashmore2015rosplan} or connect symbolic search to sampling-based geometric modules \cite{garrett2020pddlstream}. These integrations make planning deployable on real robotic systems, but they do not remove the underlying scalability limit. When many objects can participate in the task, the symbolic-geometric interface itself becomes expensive because the system must ground, sample, and validate many candidate interactions. This limitation motivates methods that reduce the object space before full symbolic search begins.

\subsection{Learning for State Abstraction and Pruning}
To motivate learned object reduction, we next review learning-based methods for state abstraction and pruning. These methods are closest to our setting because they assume the symbolic model is already available and learn guidance that makes planning over that model more efficient.

At a broad level, neuro-symbolic planning addresses scalability limits by inserting learned components into the planning pipeline. This literature follows two main directions.
First, some methods learn the symbolic interface itself. These methods lift raw observations into predicates \cite{li2023embodied, silver2023predicate, hansen2022bisimulation, li2025bilevel} or lift low-level controls into skills and operators \cite{chitnis2022learning, silver2022learning, kumar2024practice, konidaris2018skills, liu2025slap}. In robotics, such methods help make task and motion planning possible when the symbolic model is not given or incomplete in advance \cite{kumar2023learning}.
Second, another direction assumes a known symbolic model and focuses on using learning to improve planning efficiency. Early work learned heuristics or value functions to guide search \cite{arfaee2011learning, silver2016mastering}. Other work reduced grounding cost through relevance-aware instantiation, including partial grounding \cite{gnad2019learning}. More recently, learned guidance built from graph neural network representations has shown strong performance for exact discrete solvers. In combinatorial optimization, for example, Gasse et al. \cite{gasse2019exact} used a graph representation to imitate strong branching decisions in mixed-integer linear programming. The same idea has since been adapted to planning by learning object relevance. PLOI predicts which objects should be included in a simplified planning problem \cite{silver2021planning}. Chen et al. learn object-level relevance for numeric planning \cite{chen2024graph}. Flax extends this line with relaxed-plan-based recovery for missing critical objects \cite{du2026fast}.

These methods show that object pruning can cut planning time substantially, but they share a common weakness: they train the scorer from fixed offline supervision generated outside the pruned search spaces that the model induces at test time. As a result, the scorer does not learn under its own prediction errors, and exposure bias remains a central failure mode \cite{ranzato2015sequence}. Our work stays within this second direction, but replaces fixed offline supervision with adaptive pseudo-supervision generated under the scorer's current predictions.

\subsection{Neuro-Symbolic Bilevel Optimization}
We next review neuro-symbolic bilevel optimization because it provides the most natural lens for our formulation. In bilevel optimization (BLO), the upper level learns a predictor, while the lower level solves the downstream decision problem induced by that predictor \cite{amos2017optnet,poganvcic2019differentiation}. Related predict-then-optimize methods follow the same principle: they train predictions by the quality of the decisions those predictions induce rather than by offline proxy labels alone \cite{elmachtoub2022smart}. This perspective fits our setting because object-importance scores matter only through the pruned plans they produce.

The main challenge is that our lower level is a discrete symbolic planner whose search space changes with the predictions from the upper-level neural network. In continuous settings, differentiable optimization layers can often provide exact or implicit gradients \cite{amos2017optnet,chen2025ia}. Discrete settings are harder, so prior work has relied on black-box differentiation, perturbation-based surrogates, or differentiable relaxations \cite{poganvcic2019differentiation,wang2019satnet}. Yet most of this literature studies continuous control, generic combinatorial optimization, or symbolic reasoning tasks with a fixed problem instance. Our setting differs in a more specific way: pruning mistakes can change the active object set itself, which can in turn change both feasibility and the planner feedback used for learning.

Imperative learning (IL) \cite{wang2025imperative} is the closest precedent because it places neural prediction and physical or logical reasoning inside the same feedback loop. IL has shown promising results across several robotic domains \cite{chen2025ia,li2025ikap,lin2025iwalker,zhan2024imatching,guo2024imtsp,fu2024islam,yang2023iplanner,superodom}. However, applying IL to object-pruned planning is still nontrivial. Early scoring errors can remove critical objects, cause the lower-level planner to fail, and leave the upper level with unstable supervision.
Our work addresses this gap. We keep the bilevel view and the closed-loop spirit of imperative learning, but we adapt them to score-pruned task planning by learning from adaptive pseudo-supervision generated under the scorer's current predictions and by adding a stabilization strategy for lower-level search. This combination lets the scorer learn from the distribution of the task planning problems it actually creates at deployment, which directly targets the train-test mismatch of prior object-pruning methods.

\subsection{Recovery Strategies and Search Diversification}
To motivate our recovery design, we finally review prior work on recovering from search failure and diversifying symbolic search. This thread matters in our setting because online learning is only effective when the lower-level planner can still return informative plans after pruning mistakes.

In classical planning, a common strategy is \textit{plan repair}, which modifies an invalidated plan so that execution can continue without replanning from scratch \cite{fox2006plan, van2005plan}. Plan repair is useful when the planning problem itself is well specified and only the current plan has become invalid. It is less helpful when failure comes from an incomplete problem.

In parallel, another line of work improves robustness through search diversification. Portfolio planners run multiple heuristics or search configurations to reduce variance under a fixed time budget \cite{vallati2015portfolio,helmert2011fast,valenzano2012arvandherd}. Diverse planning similarly seeks multiple qualitatively different solutions \cite{xie2014type}. These methods improve coverage of the search procedure, but they assume that the planner is searching over a complete problem.

In object-pruned neuro-symbolic planning, however, failure takes a different form. When planning proceeds by iteratively adding objects \cite{silver2021planning,du2026fast}, a timeout is ambiguous: the active object set may be missing a critical object, or it may already contain too many unimportant objects. In our setting, this ambiguity matters most during bilevel optimization, because unstable lower-level search produces unstable supervision for upper-level learning. Flax \cite{du2026fast} partly addresses this with a single relaxed-plan-based repair step for recovering missing objects, but this is often insufficient to stabilize training. This gap motivates our 3R design: Repair patches local omissions, Restart rebuilds the active set from relaxed hints, and Rollback retreats to a smaller active set for conservative re-expansion.

\begin{figure}[t]
  \centering
    \includegraphics[width=1\linewidth]{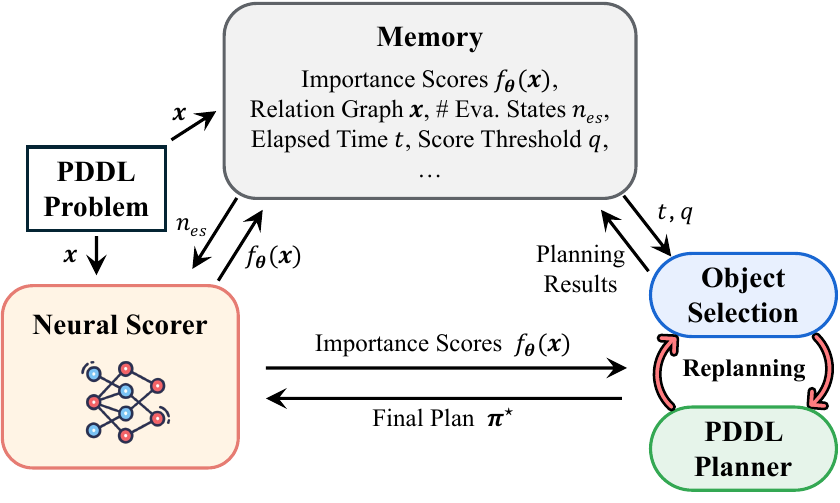}
    \vspace{-20pt}
    \caption{The training pipeline of \shortname. Given a PDDL task and its relational graph $\boldsymbol{x}$, a neural network predicts object-importance scores. The lower-level planner operates in the score-pruned search space and returns a feasible plan, providing adaptive pseudo-supervision for updating the neural scorer.}
  \label{fig:training_pipeline}
  % \vspace{-10pt}
\end{figure}

\section{Methodology}\label{sec:methodology}
We consider a PDDL planning task $\tau=\langle O,I,G\rangle$ under a fixed symbolic domain, where $O$ is the set of symbolic objects, $I$ is the initial state, and $G$ is the goal condition. Following standard PDDL terminology, an ``object'' denotes any symbolic entity, including non-physical entities such as locations.
This paper focuses on long-horizon task planning problems under complex logical constraints, where a large number of objects are present, but only a small subset is truly involved in a feasible plan. In such problems, symbolic planning over all objects is often too expensive, yet aggressive pruning is risky because removing even one critical object can make the task unsolvable.
Therefore, the practical question is not simply how to plan faster, but how to identify which objects should remain in the planning problem without destroying solvability. This is the object-pruning objective in our setting: preserve the task-relevant objects needed for a feasible plan while filtering irrelevant objects to keep symbolic search tractable. Because these two goals are coupled, we formulate planning and pruning together rather than treating object-importance learning as an isolated prediction problem.

\subsection{Neuro-Symbolic Learning Formulation}
\label{subsec:iflax}
Prior object-pruned planners typically train the scorer with fixed offline labels generated by symbolic planners in the full search space. This creates a train-test mismatch: the scorer is optimized to imitate labels from full-space plans, but at test time the planner operates inside pruned search spaces created by the scorer's own predictions, which it never encounters during training.
To close this gap, we train the scorer on plans of the pruned planning problems induced by its own predictions rather than full-space plans. Figure~\ref{fig:training_pipeline} summarizes the resulting bilevel loop based on imperative learning \cite{wang2025imperative}.
% It contains three main parts.
First, the upper-level scorer $f_{\boldsymbol{\theta}}$ maps the task representation $\boldsymbol{x}$ (translated from the PDDL problem) to object-importance scores $f_{\boldsymbol{\theta}}(\boldsymbol{x})$, where $\boldsymbol{\theta}$ denotes the scorer parameters. Second, the lower-level symbolic planner uses those scores to filter objects and search for a feasible plan $\boldsymbol{\pi^\star}$ in the corresponding pruned object space.
Third, the returned plan is converted into adaptive pseudo-supervision and fed back to update the scorer. In this way, the planner no longer serves as an offline annotation tool. Instead, it becomes part of the training iterations that teaches the scorer which objects matter for planning.
For clarity, the bilevel loop can be seen as a four-step process:
\begin{enumerate}[nosep]
\item Predict an importance score for every object in the task.
\item Run the symbolic planner on score-pruned problems within a fixed time budget.
\item Convert the returned plan into a binary pseudo-label mask by marking the used objects as positive labels.
\item Update the scorer so that its future predictions better match the objects that proved important for planning.
\end{enumerate}
We formalize this process as a bilevel optimization problem:
\begin{subequations}
\label{eq:iflax}
\begin{align}
    \quad & \min_{\boldsymbol{\theta}} \; U\big(f(\boldsymbol{\theta},\boldsymbol{x}),\boldsymbol{m}\big), \; m_i\!=\!\mathbbm{1}[o_i \in O(\boldsymbol{\pi}^\star)], \label{eq:blo-upper}\\
    \quad & \qquad \text{s.t.} \quad \boldsymbol{\pi}^\star,\boldsymbol{\nu}^\star \in \arg\min_{\boldsymbol{\pi},\ \boldsymbol{\nu}} \ L\big(\boldsymbol{x},f(\boldsymbol{\theta},\boldsymbol{x}),\boldsymbol{\pi},\boldsymbol{\nu}\big), \label{eq:blo-lower}\\
\quad & \qquad \qquad \text{s.t.} \quad \xi(M(\boldsymbol{\nu}))\le0.
\end{align}
\end{subequations}
The upper-level objective $U$ denotes the training signal used to improve the scorer: it encourages $f(\boldsymbol{\theta},\boldsymbol{x})$ to assign higher scores to objects that will appear in the final returned plan $\boldsymbol{\pi}^\star$; $O(\boldsymbol{\pi}^\star)$ denotes the set of objects that appear in the returned plan $\boldsymbol{\pi}^\star$; and $\mathbbm{1}[\cdot]$ is the indicator function, which denotes the binary pseudo-label mask $\boldsymbol{m}$.
The lower-level objective $L$ is the planner's heuristic estimate of the remaining distance to the goal from the current symbolic search node in plan $\boldsymbol{\pi}$. Under the current predicted scores $f(\boldsymbol{\theta},\boldsymbol{x})$, it favors search decisions that reduce this estimated remaining distance within the time budget;
the lower-level planner returns the final plan $\boldsymbol{\pi}^\star$ together with the final memory state $\boldsymbol{\nu}^\star$, where the memory module $M$ stores the evolving search context $\boldsymbol{\nu}$ and $\xi(M(\boldsymbol{\nu}))\!\le\!0$ is a constraint enforcing the planning time budget. 
For clarity, we next present upper-level object-importance learning, then specify the lower-level planning objective and planning algorithm, and finally introduce the bilevel optimization procedure in Section~\ref{subsec:blo}.

\subsection{Upper-Level Object-Importance Scoring}\label{subsec:scoring}
At the upper level, each task is encoded as a graph and passed to the neural scorer, which predicts the objects most likely to be relevant to a feasible plan. The goal of this stage is not to solve the task directly, but to produce object-importance scores that guide the lower-level planner toward a smaller and more relevant search space.
To this end, we first represent a PDDL task as a relational graph, then process it with a graph network $f_{\boldsymbol{\theta}}$ to predict one importance score for each object.

\subsubsection{Relational Graph Network}
Concretely, given a task instance $\tau\!=\!\langle O, I, G \rangle$, where $O$ is the object set, the initial state $I$ is the set of grounded predicates that are \textit{true} in the beginning and the goal $G$ is the set of grounded predicates that are required to be \textit{true} in the end, we represent the task as a directed multigraph $\boldsymbol{x}\!=\!(\mathcal{V},\mathcal{E})$, where $\mathcal{V}$ denotes the node set and $\mathcal{E}$ denotes the edge set.
% \subsubsection{Nodes}
The node set $\mathcal{V}$ contains one node per object in $O$, so each node stores the object-level logical information associated with one symbolic entity. The feature $\mathbf{v}_i$ for object $o_i \in O$ has three parts: a one-hot encoding of the object type, a multi-hot vector over unary predicate types whose entries indicate which predicates are \textit{true} for $o_i$ in $I$, and an analogous multi-hot vector indicating which unary predicates are required to be \textit{true} for $o_i$ in $G$.

The edge set $\mathcal{E}$ encodes relational logical constraints between objects that appear in the initial state $I$ or the goal $G$, such as spatial relations. If a grounded predicate in $I$ or $G$ involves two objects $o_i$ and $o_j$, we add a directed edge $(i,j)$ between their corresponding nodes to represent this relation. Thus, any binary relation that is explicitly stated between two objects becomes a graph connection. If a grounded predicate involves more than two objects, we convert it into pairwise directed edges among all participating objects so that the graph still records which objects jointly participate in that relation. The edge feature $\mathbf{e}_{i,j}$ is then a multi-hot vector that indicates which predicate types connect $o_i$ and $o_j$, and whether each relation comes from the initial state, the goal, or both.

To allow each object's score to depend on both its own attributes and its relations to other objects, we use an encode-process-decode graph network \cite{battaglia2018relational}. The encoder maps raw node and edge features into latent embeddings. The processor then performs $K$ rounds of message passing so each object representation can absorb information from related objects and relation edges in the graph. Notably, this design is generalizable. Because the representation is object-centric, the same graph network can naturally handle task instances with different numbers of objects. The decoder finally converts the resulting embedding of each object into a scalar importance score $p_i = [f_{\boldsymbol{\theta}}(\boldsymbol{x})]_i$.
The resulting score set $\{p_1,p_2,\dots,p_{|O|}\}$ serves as learned guidance for the lower-level planner.

\subsubsection{Upper-Level Cost}\label{subsubsec:ulo}
The upper-level cost measures the difference between the scorer's current predictions and the binary pseudo-label masks generated by the lower-level planner.
Specifically, the lower-level planner searches for a feasible plan given the current predicted scores.
Once the planner returns the plan $\boldsymbol{\pi}^\star$, we convert it into a binary pseudo-label mask $\boldsymbol{m}$. For each object $o_i$, we set $m_i\!=\!1$ if $o_i$ appears in $\boldsymbol{\pi}^\star$, and $m_i\!=\!0$ otherwise. In this sense, the planner acts as an online teacher or, more formally, an optimization-based constraint: the objects that appear in the returned plan become positive pseudo-labels for the scorer. The upper-level objective $U$ in \eqref{eq:blo-upper} is the binary cross-entropy (BCE) loss between the predicted scores $\{p_1,p_2,\dots,p_{|O|}\}$ and the mask $\boldsymbol{m}$:
\begin{equation}
U = -\frac{1}{|O|} \sum_{i=1}^{|O|} \Big( m_i \log p_i + (1 - m_i) \log (1 - p_i) \Big).
\end{equation}

\subsection{Lower-Level Planning Objective}\label{subsubsec:llo}
We now instantiate the lower-level objective $L$ in \eqref{eq:blo-lower} for the symbolic planner used in our method. The lower level maintains a memory state $\boldsymbol{\nu}$ that records the evolving search context, including the relational graph $\boldsymbol{x}$, the object-importance scores $f_{\boldsymbol{\theta}}(\boldsymbol{x})$, the temporary plan $\boldsymbol{\pi}$, the elapsed time $t$, the current score threshold $q$ that determines which objects enter the current simplified task, the number of evaluated states $n_{es}$, and the current object subset. The time budget constraint $\xi$ enforces the overall planning budget; later, in the planning pipeline, we divide that budget into an expansion phase with budget $T_1$ and a recovery phase with budget $T_2$.
Within each planning attempt, we instantiate the lower-level objective $L$ with the symbolic planning cost of Fast Downward under the \texttt{lama-first} configuration \cite{richter2010lama}, which emphasizes rapid plan discovery rather than strict plan optimality. Concretely, for a temporary plan $\boldsymbol{\pi}$, let $s(\boldsymbol{\pi})$ denote the symbolic search node reached from the initial state by following $\boldsymbol{\pi}$. Then $L$ is the heuristic estimate of the remaining distance from $s(\boldsymbol{\pi})$ to the goal. The planner minimizes this quantity with multi-queue greedy best-first search, alternating between the Fast-Forward heuristic $h_{\text{FF}}$ \cite{hoffmann2001ff} and the landmark-count heuristic $h_{\text{LM}}$ \cite{richter2008landmarks}. Let $Q_k$ be the open-list queue associated with heuristic $h_k$, where $k \in \{\text{FF}, \text{LM}\}$. At each search step, the planner pops the most promising search state $s^\star$ from the active queue:
\begin{equation}
s^\star = \arg\min_{s \in Q_k} h_k\big(s \mid \boldsymbol{x},f_{\boldsymbol{\theta}}(\boldsymbol{x}),\boldsymbol{\pi},\boldsymbol{\nu}\big).
\end{equation}
The lower-level cost at that step is
\begin{equation}
L = h_k\big(s^\star \mid \boldsymbol{x},f_{\boldsymbol{\theta}}(\boldsymbol{x}),\boldsymbol{\pi},\boldsymbol{\nu}\big).
\end{equation}
Search succeeds when $L=0$, which means the goal has been reached and the planner returns $\boldsymbol{\pi}^\star$. It fails if the task is proven unsolvable or the time budget is exhausted.

We next explain the logic of the bilevel optimization in \eqref{eq:iflax}.
Given the object-importance scores from the upper level, the lower-level planner in \eqref{eq:blo-lower} turns them into pruned planning problems and returns plans that produce adaptive pseudo-supervision. However, when those scores are still inaccurate, the planner often fails to return feasible plans due to the misleading pruning, which can destabilize the training process. 

In what follows, we first describe score-guided iterative expansion, then explain how imperfect scores create early-stage training instability and how the 3R strategy addresses it. Finally, we explain the full bilevel optimization loop.

\subsection{Score-Guided Iterative Expansion}
At the lower level, within a fixed time budget $T_1$, the planner attempts to solve pruned symbolic tasks built from the active object set, which is iteratively expanded based on the scorer's current predictions. Instead of planning over the full task $\tau$ from the start, we begin with a simplified task $\tau_{O_0}\!=\!\langle O_0,I_{O_0},G\rangle$, where $O_0 \subseteq O$ only contains the objects that appear in the goal $G$. At iteration $N$, the planner attempts to solve the simplified task $\tau_{O_N}\!=\!\langle O_N,I_{O_N},G\rangle$ with
$
O_N\!=\!O_0 \cup \{o_i\!\in\!O \mid p_i\!\ge\!q_N\}.
$
If the simplified task is proved unsolvable quickly, we lower the score threshold $q$ with decay factor $\gamma$, namely $q_{N+1}\!=\!\gamma q_N$, and add the next batch of high-scoring objects $O_{\text{batch}}^{N}\!=\!O_{N+1}\!\setminus\!O_N$. In this way, the search grows in the order suggested by the scorer rather than committing immediately to the full combinatorial space.
If iterative expansion returns a feasible plan, the lower level terminates immediately. Otherwise, if the time budget runs out, the planner treats the failure as a signal that the current score-pruned search space needs recovery.

\begin{algorithm}[t]
  \caption{\shortname Planning Pipeline}
  \label{alg:iflax_planning}
  \SetKwInOut{Input}{Input}\SetKwInOut{Output}{Output}
  \SetKwBlock{Parallel}{Parallel execute with time budget $T_2$:}{}
  \Input{Task $\tau=\langle O, I, G \rangle$ and its relational graph $\boldsymbol{x}$; Neural Scorer $f_{\boldsymbol{\theta}}$; Expansion Budget $T_1$; Recovery Budget $T_2$; Initial Threshold $q_{\max}$; Decay Factor $\gamma$}
  \Output{Solution $\pi^\star$ (or \textit{\textbf{fail}})}
  \BlankLine
  \tcp*[l]{Score-Guided Iterative Expansion}
  $S \gets f_{\boldsymbol{\theta}}(\boldsymbol{x})$ \tcp*{importance scores $p_i \in (0,1)$}
  $O_0 \gets \{\,o \in O \mid o \text{ appears in } G\,\}$\;
  $q \gets q_{\max}$;\quad $N \gets 0$;\quad $t \gets 0$\;
  \While{$t < T_1$}{
      $O_N \gets O_0 \cup \{\,o_i \in O \mid p_i \ge q\,\}$\;
      $\pi \gets \textsc{Plan}(\tau_{O_N},\; T_1 - t)$\;
      \If{$\pi \neq \textbf{fail}$}{
          \Return $\pi$\;
      }
      \If{\text{Timeout on} $\tau_{O_N}$}{
          \textbf{break} \tcp*{Stuck, trigger recovery}
      }
      $q\!\gets\!\gamma q$;\quad $N\!\gets\!N+1$ \tcp*{Expand obj. set}
  }
  \BlankLine
  \tcp*[l]{3R Strategy}
  \If{$\pi = \textbf{fail}$}{
      \Parallel{
          \tcp{Thread 1: Repair}
          $\tilde{\pi} \gets \textsc{Plan}(\textsc{Relax}(\tau),\; T_2)$\;
          $\tilde{O} \gets \{\,o \in O \mid o \text{ appears in } \tilde{\pi}\,\}$\;
          $O_{\text{Repair}} \gets \textsc{Comp}(O_N \cup (\tilde{O} \cap O))$\;
          $\pi_{\text{Repair}} \gets \textsc{Plan}(\tau_{O_{\text{Repair}}},\; T_2)$\;
          \BlankLine
          \tcp{Thread 2: Restart}
          $O_{\text{Restart}} \gets \textsc{Comp}(O_0 \cup (\tilde{O} \cap O))$\;
          $\pi_{\text{Restart}} \gets \textsc{IterativeExp}(O_{\text{Restart}}, S, T_2)$\;
          \BlankLine
          \tcp{Thread 3: Rollback}
          % $O_{\text{Rollback}} \gets O_{N-1}$\;
          $\pi_{\text{Rollback}} \gets \textsc{IncrementalExp}(O_{N-1}, S, T_2)$\;
      }
      $\pi^\star\!\gets\!\text{first valid plan from } \{\pi_{\text{Repair}},\!\pi_{\text{Restart}},\!\pi_{\text{Rollback}}\}$\;
      \Return $\pi^\star$\;
  }
  \BlankLine
  \vspace{-0.15em}
  \hrule
  \vspace{0.15em}
  \small
  \textbf{Auxiliary procedures}\\[2pt]
  \Indp
  $\textsc{Plan}(\tau,\Delta t)$: runs the symbolic planner on task $\tau$ with time budget $\Delta t$; returns $\pi$ or \textit{\textbf{fail}}\\
  $\textsc{Relax}(\tau)$: returns a relaxed task derived from $\tau$ by applying predefined domain-specific relaxation rules\\
  $\textsc{Comp}(\mathcal{O})$: returns the closure of object set $\mathcal{O}$ under domain-specific complementary rules\\
  $\textsc{IterativeExp}(\mathcal{O}, S, \Delta t)$: runs score-guided iterative expansion logic using $\mathcal{O}$ as the base set $O_0$\\
  $\textsc{IncrementalExp}(\mathcal{O}, S, \Delta t)$: resumes expansion but adds objects one-by-one in descending score order\\
  \Indm
\end{algorithm}

\subsection{3R Strategy}\label{subsubsec:3r}
The 3R strategy is what makes online bilevel training practical: it stabilizes score-guided iterative expansion and provides adaptive pseudo-supervision even when object scores are still imperfect. Without reliable recovery, early search failures can prevent the lower-level planner from returning feasible solutions and break the learning signal.
Specifically, during early epochs, the scores can be unreliable, so iterative expansion may get stuck and fail to return a feasible plan. If score-guided iterative expansion and recovery both fail within the allocated budgets, we discard this task for the current epoch. The task then provides no adaptive pseudo-supervision in this pass, but it is revisited in later epochs after the scorer has improved. If such failures occur on many tasks, the scorer is updated from only a small and biased subset of successful cases, which can make its predictions less reliable in the next epoch. The learning loop can then enter a vicious cycle: poor scores lead to plan failures, plan failures leave fewer successful tasks to supervise the scorer, and the resulting biased updates further degrade score quality in the next epoch of training.

Concretely, search failure occurs when the planner exhausts its time budget $T_1$ on the active object set $O_N$ without finding a plan or proving the simplified task unsolvable.
In our experiments, we repeatedly observe three qualitatively different reasons for such failures. First, the scorer may omit a critical object.
Second, the active set may become bloated with unimportant objects from earlier iterations, which makes heuristic search spend effort on many irrelevant interactions. Third, the current active set may already be close to sufficient, but the next update can add too many objects at once and trigger a sharp jump in search cost. These three failures require different recovery strategies: missing objects call for adding the right context without discarding useful progress; bloated active sets call for removing clutter rather than appending even more objects; over expansion calls for a more cautious expansion, not a full reset. We therefore terminate an early iterative expansion when the planner gets stuck and launch three parallel recovery branches, each with time budget $T_2$. Their roles are complementary: \textit{Repair} addresses local incompleteness, \textit{Restart} addresses global search-space pollution, and \textit{Rollback} addresses abrupt over-expansion. \algref{alg:iflax_planning} summarizes the planning procedure with score-guided iterative expansion and parallel 3R recovery strategy.

\subsubsection{Repair}
First, Repair targets critical object omissions when the current active set is mostly correct but incomplete.
Inspired by Flax \cite{du2026fast}, we use a relaxed plan to recover objects that may be missing from the current active set. But rather than introducing Repair as the only recovery method as Flax does, we treat it as one branch of our broader 3R recovery framework. Because this local patch can address missing objects but cannot remove accumulated irrelevant ones or revert an overly aggressive expansion step. 
The main advantage of this design is that it is general and lightweight. Relaxation rules only need to simplify a domain while preserving a rough task skeleton, and complementary rules only need to restore tightly coupled symbolic context around the objects suggested by that relaxed sketch. In many object-centric domains, both kinds of rules are easy to specify from basic domain structure rather than from case-specific engineering. 
Concretely, we solve a relaxed version of the full task to obtain a rough plan $\tilde{\pi}$, extract the involved objects $\tilde{O}$, patch $(\tilde{O}\cap O)$ into the current active set $O_N$, and finally apply complementary rules to get the resulting set $O_{\text{Repair}}$. Next, we will introduce these rules in detail.

In \algref{alg:iflax_planning}, \textsc{Relax} means that we use relaxation rules to temporarily simplify some constraints so the planner can quickly produce a rough plan. The design principle is simple: remove or weaken constraints that make search hard, but keep enough causal structure for the relaxed plan to still indicate which objects are likely to matter in the original task. For example, in \textit{MazeNamo} \cite{du2026fast}, a grid rearrangement domain where the robot must reach the goal location by moving obstacles, the relaxation rule can be interpreted as \textit{removing all light obstacles}. This produces an easier maze that often preserves the corridor structure and reveals which heavy boxes and locations still matter for reaching the goal. In \textit{LogisticsPlus} (a harder version of \textit{Logistics} \cite{mcdermott20001998} we proposed), a transport domain with cities, vehicles, locked hubs, and key-gated routes, one relaxation rule can be interpreted as \textit{unlocking all locked hubs}, which relaxes the rigor transport dependencies without damaging the original task structure.

We then apply complementary rules \textsc{Comp} to restore related objects that may be absent from the simplified task. The design principle here is equally general and simple: when the relaxed plan points to an object, we add the tightly coupled symbolic context required to make that object usable in the original task. For example, in \textit{MazeNamo}, where obstacles occupy grid locations and rearrangement actions depend on those local occupancies, an obstacle and the location it occupies should be added together. If Repair adds a heavy obstacle to the active set, \textsc{Comp} also adds the location where that obstacle initially sits. Conversely, if Repair adds a location because the relaxed plan needs to route through or manipulate that place, \textsc{Comp} also adds the obstacle that initially occupies that location. In this way, the repaired task preserves the local item-location dependency, so the symbolic state remains coherent. The resulting set $O_{\text{Repair}}$ defines a patched simplified task $\tau_{O_{\text{Repair}}}$ for replanning. Repair is therefore the right branch when the planner mainly needs a precise local correction and the rest of $O_N$ is still worth preserving.

\begin{figure}[t]
  \centering
    \includegraphics[width=1\linewidth]{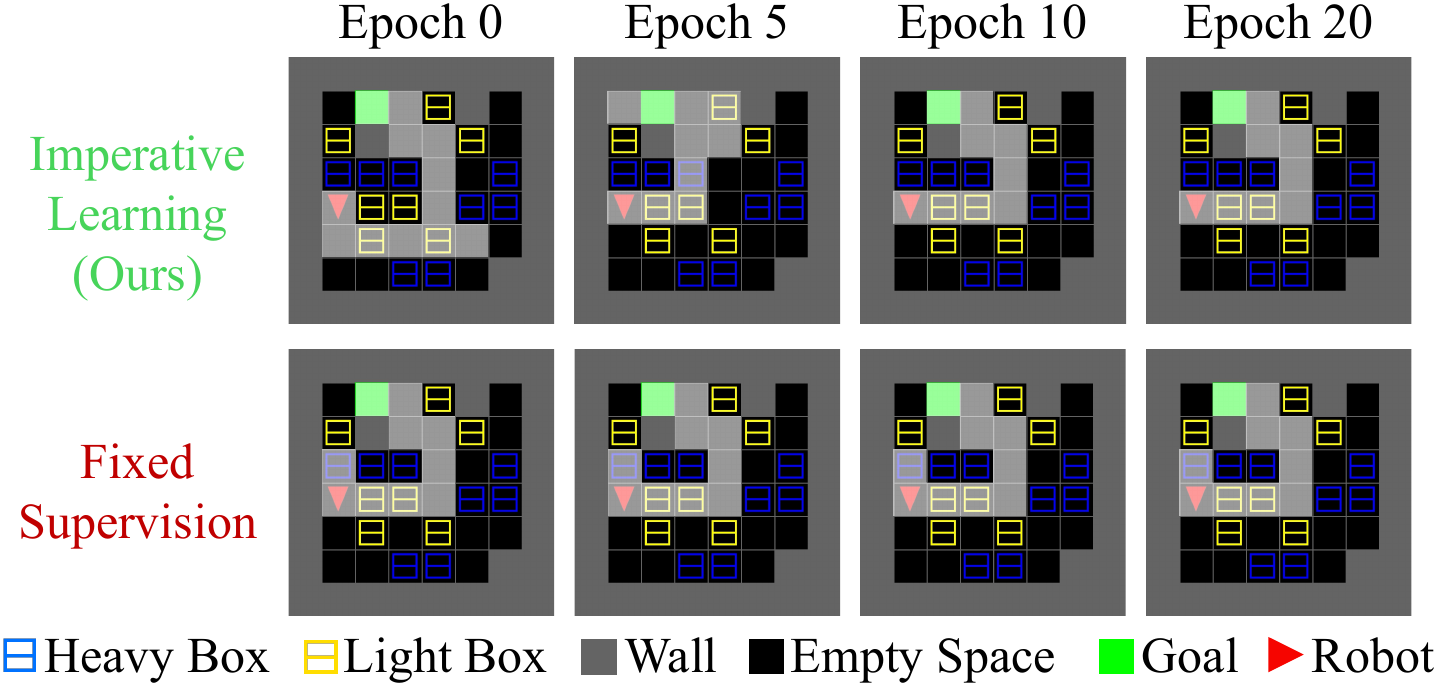}
    \caption{Evolution of the supervision plan during training in \textit{MazeNamo}. Under imperative learning (top), the plan changes with the scorer's predictions. Under fixed supervision (bottom), the plan remains unchanged across training.}
  \label{fig:supervision_plan_vis}
  % \vspace{-10pt}
\end{figure}

\begin{figure*}[t]
  \centering
    \includegraphics[width=1\linewidth]{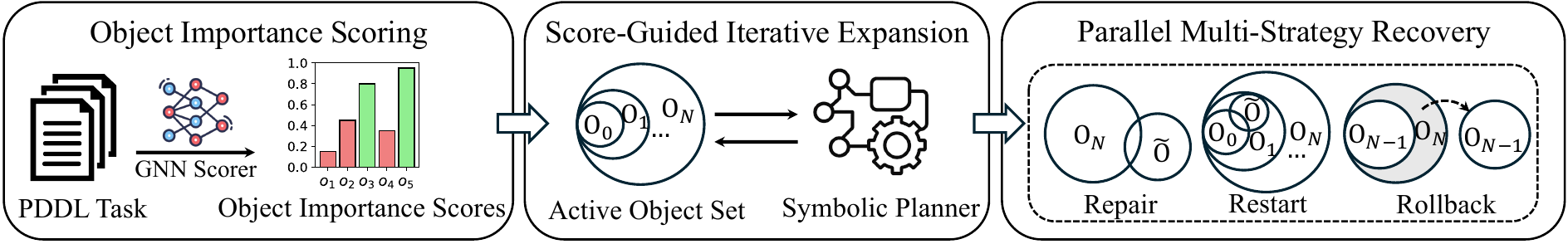}
    \caption{Planning pipeline of \shortname. The scorer predicts object-importance scores, the planner performs score-guided iterative expansion over active object sets, and, if search gets stuck, it launches parallel \textit{Repair}, \textit{Restart}, and \textit{Rollback} recovery.}
  \label{fig:planning_pipeline}
  % \vspace{-10pt}
\end{figure*}

\subsubsection{Restart}
Second, Restart targets search-space bloat caused by too many unimportant objects being assigned high scores. This failure mode is fundamentally different from local omission fixed by Repair. If $O_N$ is already polluted, patching more objects into it usually makes the search harder rather than easier, so Repair alone cannot solve the problem. Restart therefore reuses the same rules in Repair for a different purpose. Instead of patching the current set, it discards that set and rebuilds a cleaner one. We first solve the relaxed problem to obtain a rough plan $\tilde{\pi}$, then extract the objects $\tilde{O}$ that appear in that plan. We next apply the same complementary rules to recover the related objects required for a coherent symbolic state. This produces a new base set $O_{\text{Restart}}$, which is typically much smaller and cleaner than the bloated set $O_N$. We then use $O_{\text{Restart}}$ as the new initial set for score-guided iterative expansion and continue expanding it until a feasible plan is found or the recovery budget $T_2$ is exhausted. The key novelty is that Restart changes the recovery from local patching to full search-space re-initialization. That distinction matters because many training failures are caused by accumulated unimportant objects, an unsolved case for prior single-repair methods. Whenever a relaxed task sketch is more trustworthy than the bloated active set, rebuilding from that sketch is preferable.

\subsubsection{Rollback}
Third, Rollback handles over expansion, where the number of newly added objects in a single expansion is large and leads to a timeout. Here, the issue is neither a missing critical object nor a badly polluted active set. Instead, the current search direction is often correct, but the latest update of the active object set adds too many objects at once and turns a tractable problem into an intractable one. In that case, repairing with more objects misses the source of failure, while restarting from scratch is unnecessarily disruptive. 
One could replace threshold-based expansion from the beginning with fixed-size expansion, which adds a fixed number of top-ranked objects at each step, but that is not strictly better: when the scores are already accurate and many important objects should enter together, this strategy can waste planning attempts on several unsolvable simplified tasks and exhaust the time budget before reaching the right active set. 
Rollback avoids this global tradeoff. We keep the efficient coarse expansion when it works, and invoke fine-grained one-by-one expansion when failure recovery is needed. 
Concretely, we step back from the current active set $O_N$ to $O_{N-1}$ and continue expansion one object at a time instead of adding a whole batch. For example, if lowering the threshold adds several medium-score obstacles and locations at once and leads to the planner getting stuck, Rollback returns to the last manageable active set and reintroduces those objects gradually. This finer-grained search stays near the ``boundary'' of tractability without triggering another combinatorial jump. The key idea behind Rollback is that some failures arise not from choosing the wrong objects, but from introducing the right objects too abruptly, in which case this fine-grained one-by-one expansion is more appropriate than resetting the search entirely.

For training stability, the three parallel branches use the same time budget $T_2$, and if multiple branches return feasible plans, we keep the one with the fewest evaluated states $n_{es}$, which directly reflects the size of search space. This gives a more stable supervision signal because it depends on search effort rather than unstable running time, which is often affected by incidental fluctuations in the hardware status. 
More importantly, the 3R design is general because its three branches address broad search failure modes, missing critical objects, search-space bloat, and over expansion, rather than domain-specific cases.
In this way, 3R is not an add-on after the bilevel formulation is chosen. Rather, it makes the online formulation trainable in practice by covering these failure modes and restoring reliable lower-level search when early-stage pruning would otherwise break the learning loop.

\subsection{Bilevel Optimization}\label{subsec:blo}
Together, score-guided iterative expansion and 3R recovery strategy define the lower-level solver in \eqref{eq:blo-lower}. In each training iteration, the scorer first predicts the object-importance scores $f_{\boldsymbol{\theta}}(\boldsymbol{x})$. The planner then runs the lower-level pipeline in \algref{alg:iflax_planning} and returns a feasible plan $\boldsymbol{\pi}^\star$, which is converted into a binary pseudo-label mask by marking the objects that appear in the plan as positive and the rest as negative. The upper level then applies the BCE loss to update the scorer for the next training iteration.
This adaptive pseudo-supervision closes the bilevel loop and overcomes the \textit{exposure bias} in prior methods. As shown in \fref{fig:supervision_plan_vis}, the planner-generated plan, and therefore the pseudo-labels in \shortname, evolve with the training process, whereas under prior methods, the supervision plan remains frozen during training. Because the scorer is trained on plans found inside its own score-pruned search spaces, the training signal is aligned more closely with the distribution of the planning tasks it will face at deployment.

\subsection{Deployment-Time Planning Pipeline}\label{subsec:deployment}
After training, the scorer parameters are fixed. Deployment then uses the frozen scorer together with the symbolic planning pipeline in \fref{fig:planning_pipeline}. For a given task, the scorer is evaluated once to assign an importance score to each object. The planner then uses these scores to run score-guided iterative expansion, starting from the goal-relevant objects and progressively adding higher-scoring objects as needed. If search gets stuck, the planner launches the three 3R recovery branches, namely \textit{Repair}, \textit{Restart}, and \textit{Rollback}, in parallel under the remaining time budget. Unlike training, where we keep the successful branch with the fewest evaluated states to stabilize supervision, deployment stops as soon as any branch returns a feasible plan and uses that first successful branch as the final solution.
This parallel early-stop design is both reliable and fast. It is reliable because the three branches target the dominant failure modes identified in \sref{subsubsec:3r}, so the recovery system covers most failures encountered at deployment. It is fast because the planner does not need to diagnose the failure mode or try strategies one after another; instead, the matching branch can succeed immediately while the others are terminated.

This deployment pipeline is simpler than the training loop because it removes the pseudo-supervision path: there is no upper-level loss, no online label generation, and no parameter update. As a result, the neural module adds only a small fixed inference cost, while significantly reducing the search cost spent on symbolic planning. In practice, this separation makes the method easy to integrate into a robotic system once the scene has been converted into the similar PDDL representation and domain rules used during training. The main deployment benefit of \shortname is therefore not only better object scoring, but also more efficient use of the planning time budget through progressive search-space expansion and targeted recovery.

\begin{table*}[!t]
    \caption{Planning performance on \textit{MazeNamo}. ``FR'' denotes Failure Rate, ``WPT'' denotes Weighted Planning Time, and the last two columns report relative gains over Flax. Lower is better for both metrics. Best results are bolded and underlined.}
    \label{tab:main_results_maze}
    \centering
    \begin{tabular}{@{}C{0.07\linewidth}@{}C{0.07\linewidth}@{}*{10}{C{0.085\linewidth}@{}}}
        \toprule
        \multirow{2}{*}{Map Size} & \multirow{2}{*}{Level} & \multicolumn{2}{c}{LAMA \cite{richter2010lama}} & \multicolumn{2}{c}{PLOI \cite{silver2021planning}} & \multicolumn{2}{c}{Flax \cite{du2026fast}} & \multicolumn{2}{c}{\shortname\ (Ours)} & \multicolumn{2}{c}{\makecell[c]{Improv. vs Flax}} \\
        & & \makecell[c]{FR $\downarrow$} & \makecell[c]{WPT (s) $\downarrow$} & \makecell[c]{FR $\downarrow$} & \makecell[c]{WPT (s) $\downarrow$} & \makecell[c]{FR $\downarrow$} & \makecell[c]{WPT (s) $\downarrow$} & \makecell[c]{FR $\downarrow$} & \makecell[c]{WPT (s) $\downarrow$} & \makecell[c]{FR $\downarrow$} & \makecell[c]{WPT (s) $\downarrow$} \\
        \midrule
        \multirow{4}{*}{10$\times$10} & Easy & \underline{\textbf{0.000}} & 0.94 & 0.050 & 0.75 & 0.020 & 0.62 & \underline{\textbf{0.000}} & \underline{\textbf{0.47}} & -100.0\% & -23.70\% \\
        & Medium & \underline{\textbf{0.000}} & 1.99 & 0.096 & 0.97 & 0.024 & 0.79 & 0.002 & \underline{\textbf{0.60}} & -91.67\% & -24.56\% \\
        & Hard & 0.250 & 4.27 & 0.251 & 1.90 & 0.104 & 1.90 & \underline{\textbf{0.035}} & \underline{\textbf{1.04}} & -66.35\% & -45.23\% \\
        & Expert & 1.000 & 5.00 & 0.510 & 3.12 & 0.379 & 3.03 & \underline{\textbf{0.060}} & \underline{\textbf{1.12}} & -84.17\% & -62.95\% \\
        \cmidrule(lr){1-12}
        \multirow{4}{*}{12$\times$12} & Easy & \underline{\textbf{0.000}} & 4.35 & 0.107 & 3.18 & 0.023 & 1.92 & \underline{\textbf{0.000}} & \underline{\textbf{1.20}} & -100.0\% & -37.45\% \\
        & Medium & \underline{\textbf{0.000}} & 8.86 & 0.158 & 4.49 & 0.031 & 2.98 & 0.002 & \underline{\textbf{1.66}} & -93.55\% & -44.40\% \\
        & Hard & 0.520 & 18.85 & 0.476 & 10.92 & 0.170 & 8.09 & \underline{\textbf{0.076}} & \underline{\textbf{3.73}} & -55.29\% & -53.82\% \\
        & Expert & 1.000 & 20.00 & 0.471 & 10.88 & 0.364 & 10.26 & \underline{\textbf{0.054}} & \underline{\textbf{3.10}} & -85.16\% & -69.78\% \\
        \cmidrule(lr){1-12}
        \multirow{4}{*}{15$\times$15} & Easy & \underline{\textbf{0.000}} & 21.46 & 0.142 & 9.28 & 0.026 & 7.49 & \underline{\textbf{0.000}} & \underline{\textbf{3.60}} & -100.0\% & -51.90\% \\
        & Medium & 0.370 & 36.35 & 0.213 & 12.63 & 0.065 & 12.30 & \underline{\textbf{0.019}} & \underline{\textbf{4.83}} & -70.77\% & -60.75\% \\
        & Hard & 0.920 & 39.78 & 0.253 & 14.20 & 0.079 & 13.51 & \underline{\textbf{0.019}} & \underline{\textbf{4.70}} & -75.95\% & -65.21\% \\
        & Expert & 1.000 & 40.00 & 0.526 & 24.40 & 0.393 & 23.77 & \underline{\textbf{0.068}} & \underline{\textbf{7.77}} & -82.70\% & -67.33\% \\
        \cmidrule(lr){1-12}
        \multicolumn{2}{c}{\textit{Average}} & 0.422 & 70.68\% & 0.271 & 36.13\% & 0.140 & 32.14\% & \underline{\textbf{0.028}} & \underline{\textbf{13.78\%}} & -80.04\% & -57.14\% \\
        \bottomrule
    \end{tabular}
\end{table*}

\section{Experimental Evaluation}
We evaluate \shortname from four aspects. First, we test whether it improves overall planning robustness and efficiency on three challenging PDDL benchmarks with large object sets.
Second, we analyze how bilevel optimization addresses exposure bias relative to fixed offline supervision. Third, we evaluate how the 3R recovery strategy stabilizes training and improves test-time performance. 
Fourth, we test whether the pipeline supports complete execution in simulation and the real world.

\subsection{Experimental Setup}
% \subsubsection{Baselines and Evaluation Metrics}
We evaluate \shortname on three benchmarks, \textit{MazeNamo} \cite{du2026fast}, \textit{SokoMindPlus} \cite{sokomindplus}, \textit{LogisticsPlus} (a more challenging version of \cite{mcdermott20001998}), which aim to test different capabilities in task planning: dense obstacle rearrangement, long sequential action dependencies, and transport scheduling over graph-structured cities.
We compare \shortname against three baselines: LAMA \cite{richter2010lama}, PLOI \cite{silver2021planning}, and Flax \cite{du2026fast}.
To ensure a fair comparison, all neural methods use the same network architecture and are trained and tested on the same machine with an AMD Ryzen 9 7950X CPU and an NVIDIA GeForce RTX 4090 GPU.
The PLOI and Flax models are trained for 300 epochs in all domains.
For \shortname, we train for 20 epochs in \textit{MazeNamo}, 10 epochs in \textit{SokoMindPlus}, and 50 epochs in \textit{LogisticsPlus}.

\paragraph*{Metrics}
We report two metrics: \textit{Failure Rate} (FR), the fraction of instances unsolved within the time budget, and \textit{Weighted Planning Time} (WPT), the average planning time where unsolved instances contribute the full time budget. All results are averaged over 10 runs with random seeds.

\subsection{Performance on MazeNamo}
To comprehensively test whether \shortname can identify a small set of task-relevant objects inside a large problem, we adopt \textit{MazeNamo} \cite{du2026fast} as our main benchmark.
Each task is an $n\!\times\!n$ grid with boundary walls and randomly generated interior walls, heavy obstacles, light obstacles, free cells, a robot, and a goal location. The challenge is not only the number of objects, but also the complex interaction constraints: the robot (1) must reason about orientation, (2) can move only into empty cells, (3) can push movable obstacles only when the destination cell is empty, (4) can pick up only light obstacles, and (5) can stack/unstack light obstacles subject to the \texttt{upon}, \texttt{onGround}, and \texttt{clear} predicates. These constraints make the search space extremely large because many obstacles are movable, but only a few lie on the dependency chain of actions required to reach the goal.

The object scale is also substantial. We train the models on 200 easy $8\!\times\!8$ instances but evaluate on instances with map sizes $10\!\times\!10$, $12\!\times\!12$, and $15\!\times\!15$, each split into \textit{easy}, \textit{medium}, \textit{hard}, and \textit{expert} levels.
% by the full-space LAMA runtime. 
In the hardest $15\!\times\!15$ expert level, each task contains about 360 symbolic objects overall, which is extremely challenging for conventional planners.
This is precisely the regime where object pruning matters, because adding or removing only a few obstacles can change whether the planner sees a short corridor-opening solution or a much larger irrelevant search tree.
The evaluation time limits are 5s, 20s, and 40s for three map sizes respectively.

The results in \tref{tab:main_results_maze} show that \shortname remains consistently ahead across all 12 evaluated settings. Relative to Flax, it reduces the average FR by 80.04\% and the average WPT by 57.14\%. The margin is especially large on the hardest $15\!\times\!15$ expert tasks, where FR drops from 39.3\% to 6.8\% and WPT drops from 23.77s to 7.77s. This is strong evidence that \shortname improves the capability that matters most in this challenging long-horizon benchmark: finding the few rearrangement-critical obstacles early enough to keep the grounded problem small without removing objects that later become necessary.

\subsection{Performance on SokoMindPlus}
We further test \shortname on \textit{SokoMindPlus} \cite{sokomindplus} to evaluate whether it can identify the objects that determine feasibility in long, irreversible action sequences. This domain is a harder variant of Sokoban \cite{culberson1997sokoban}. Only a small subset of objects appears in the goal, where specific boxes should be moved to designated places, while many other objects can still determine whether those goals remain reachable. Because the planner can push boxes but cannot pull them back, an early mistake can make a future route or goal permanently infeasible. In contrast to \textit{MazeNamo}, which focuses more on heterogeneous manipulation affordances, \textit{SokoMindPlus} focuses more on coupled push dependencies among many similar objects.

We train on 200 easy $10\!\times\!10$ instances and evaluate on three test sets: $15\!\times\!15$ (10s budget), $18\!\times\!18$ (40s budget), and $20\!\times\!20$ (60s budget). The hardest $20\!\times\!20$ set contains around 450 symbolic objects overall. Although only a few boxes appear in the goal, the planner must still reason through a much larger clutter field, because many non-goal boxes can become temporary blockers or permanent dead-end obstacles.

\begin{table*}[!t]
    \caption{Planning performance on \textit{SokoMindPlus}.}
    \label{tab:main_results_soko}
    \centering
    \begin{tabular}{@{}C{0.08\linewidth}@{}*{10}{C{0.09\linewidth}@{}}}
        \toprule
        \multirow{2}{*}{Map Size} & \multicolumn{2}{c}{LAMA \cite{richter2010lama}} & \multicolumn{2}{c}{PLOI \cite{silver2021planning}} & \multicolumn{2}{c}{Flax \cite{du2026fast}} & \multicolumn{2}{c}{\shortname\ (Ours)} & \multicolumn{2}{c}{\makecell[c]{Improv. vs Flax}} \\
        & \makecell[c]{FR $\downarrow$} & \makecell[c]{WPT (s) $\downarrow$} & \makecell[c]{FR $\downarrow$} & \makecell[c]{WPT (s) $\downarrow$} & \makecell[c]{FR $\downarrow$} & \makecell[c]{WPT (s) $\downarrow$} & \makecell[c]{FR $\downarrow$} & \makecell[c]{WPT (s) $\downarrow$} & \makecell[c]{FR $\downarrow$} & \makecell[c]{WPT (s) $\downarrow$} \\
        \midrule
        15$\times$15 & 1.000 & 10.00 & 0.696 & 7.75 & 0.617 & 7.45 & \underline{\textbf{0.369}} & \underline{\textbf{5.63}} & -40.19\% & -24.42\% \\
        18$\times$18 & 1.000 & 40.00 & 0.738 & 31.89 & 0.615 & 29.02 & \underline{\textbf{0.375}} & \underline{\textbf{21.57}} & -39.02\% & -25.69\% \\
        20$\times$20 & 1.000 & 60.00 & 0.634 & 42.55 & 0.528 & 39.07 & \underline{\textbf{0.314}} & \underline{\textbf{27.83}} & -40.53\% & -28.78\% \\
        \cmidrule(lr){1-11}
        \textit{Average} & 1.000 & 100.0\% & 0.689 & 76.03\% & 0.587 & 70.72\% & \underline{\textbf{0.353}} & \underline{\textbf{52.20\%}} & -39.89\% & -26.19\% \\
        \bottomrule
    \end{tabular}
\end{table*}

\begin{table*}[!t]
    \caption{Planning performance on \textit{LogisticsPlus}.}
    \label{tab:main_results_logistics}
    \centering
    \begin{tabular}{@{}C{0.08\linewidth}@{}*{10}{C{0.09\linewidth}@{}}}
        \toprule
        \multirow{2}{*}{Map Size} & \multicolumn{2}{c}{LAMA \cite{richter2010lama}} & \multicolumn{2}{c}{PLOI \cite{silver2021planning}} & \multicolumn{2}{c}{Flax \cite{du2026fast}} & \multicolumn{2}{c}{\shortname\ (Ours)} & \multicolumn{2}{c}{\makecell[c]{Improv. vs Flax}} \\
        & \makecell[c]{FR $\downarrow$} & \makecell[c]{WPT (s) $\downarrow$} & \makecell[c]{FR $\downarrow$} & \makecell[c]{WPT (s) $\downarrow$} & \makecell[c]{FR $\downarrow$} & \makecell[c]{WPT (s) $\downarrow$} & \makecell[c]{FR $\downarrow$} & \makecell[c]{WPT (s) $\downarrow$} & \makecell[c]{FR $\downarrow$} & \makecell[c]{WPT (s) $\downarrow$} \\
        \midrule
        L50 & 1.000 & 30.00 & 0.402 & 19.74 & 0.369 & 20.52 & \underline{\textbf{0.109}} & \underline{\textbf{13.08}} & -70.46\% & -36.23\% \\
        \cmidrule(lr){1-11}
        \textit{Average} & 1.000 & 100.0\% & 0.402 & 65.80\% & 0.369 & 68.40\% & \underline{\textbf{0.109}} & \underline{\textbf{43.60\%}} & -70.46\% & -36.23\% \\
        \bottomrule
    \end{tabular}
\end{table*}

The results in \tref{tab:main_results_soko} show that \shortname is best on all three test sizes. Averaged over the benchmark, it lowers FR from 58.7\% to 35.3\% and WPT from 70.72\% to 52.20\% relative to Flax. This is obvious evidence that the \shortname can preserve the causal object set needed for solvability, even when most objects are similar and early mistakes are hard to reverse.

\subsection{Performance on LogisticsPlus}
To test the capability for reasoning over symbolic transport dependencies and resource constraints, we propose the \textit{LogisticsPlus} benchmark, which is built on the classic Logistics benchmark \cite{mcdermott20001998}.
The classic domain asks a planner to move packages between locations using trucks and airplanes, where all locations within the same city are directly connected, vehicles have unlimited carrying capacity, and packages do not need to satisfy stacking or loading constraints.

Specifically, we increase its difficulty by introducing directed road and air graphs, unit-capacity trucks and airplanes, stackable packages, ground-occupancy constraints, locked hubs, switch panels, and key packages. 
Their task constraints are complex: (1) A package can be loaded only when it is clear, (2) unloading to the ground is allowed only when that location is ground-free, (3) trucks cannot enter locked hubs, and (4) some routes become usable only after a key package is delivered to the correct switch panel. Because these dependencies interact across cities and locations, this benchmark is very different from the two earlier ones.

There is also a large train-test gap in this benchmark.
Specifically, the training set contains 200 small instances with 2 cities, 3--5 locations per city, and 0--1 locked hub, and we evaluate on 20 larger instances with 5 cities, 8--12 locations per city, multiple locked hubs, and a 30s time budget. 
However, the test set contains 42--58 locations, 33--48 packages, 5 trucks, 2 airplanes, and 1--4 locked hubs, for around 100 symbolic objects overall. 
This large train-test gap in object count makes \textit{LogisticsPlus} a strong test of difficulty and generalization.
Additionally, the raw object count is lower than the earlier benchmarks, but each object is involved in much longer task plans, so deciding which objects to prune is very difficult.

The results in \tref{tab:main_results_logistics} show that \shortname reduces FR from 36.9\% to 10.9\% and WPT from 20.52s to 13.08s relative to Flax, which proves that the same training and recovery principles are not tied to geometric obstacle domains. They also improve planning when the main challenge is symbolic coupling among routes, vehicles, stacks, locks, and keys.

\begin{figure*}[t]
    \centering
    \subfloat[{Solvability Within Limited Time}]{%
        \begin{overpic}[width=0.32\linewidth]{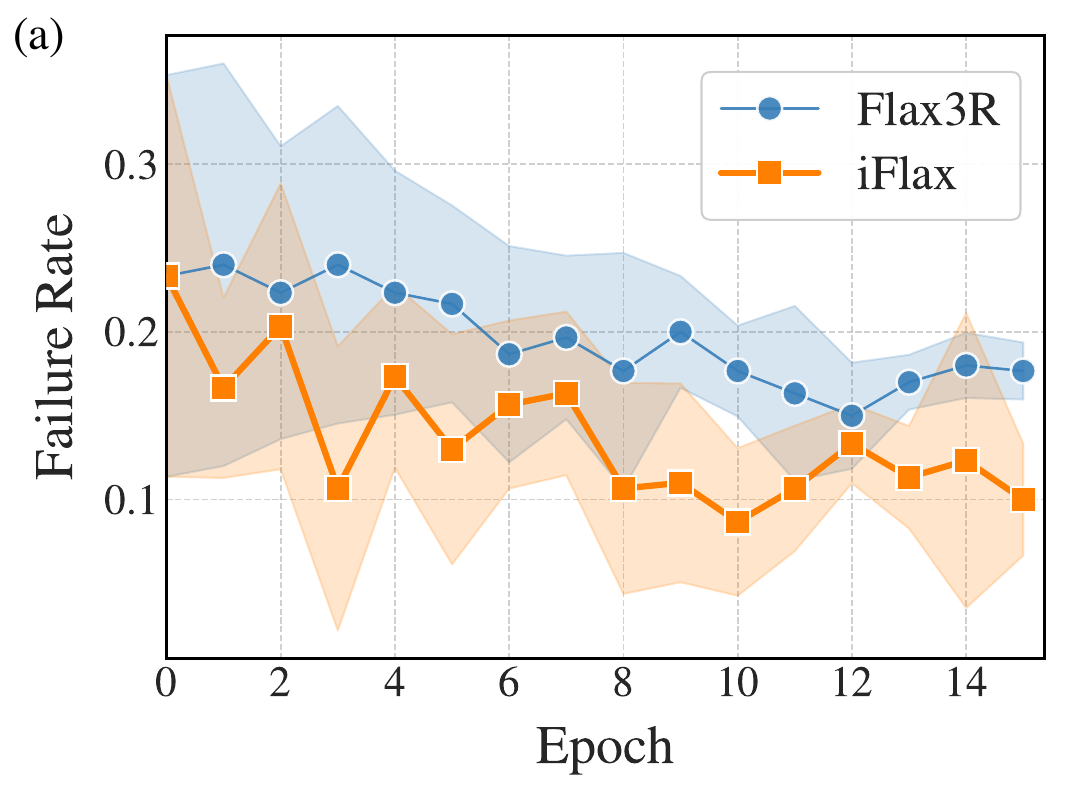}
        \end{overpic}%
        \label{fig:val_fr}%
    }
    \hfill
    \subfloat[{Identification of Important Objects}]{%
        \begin{overpic}[width=0.32\linewidth]{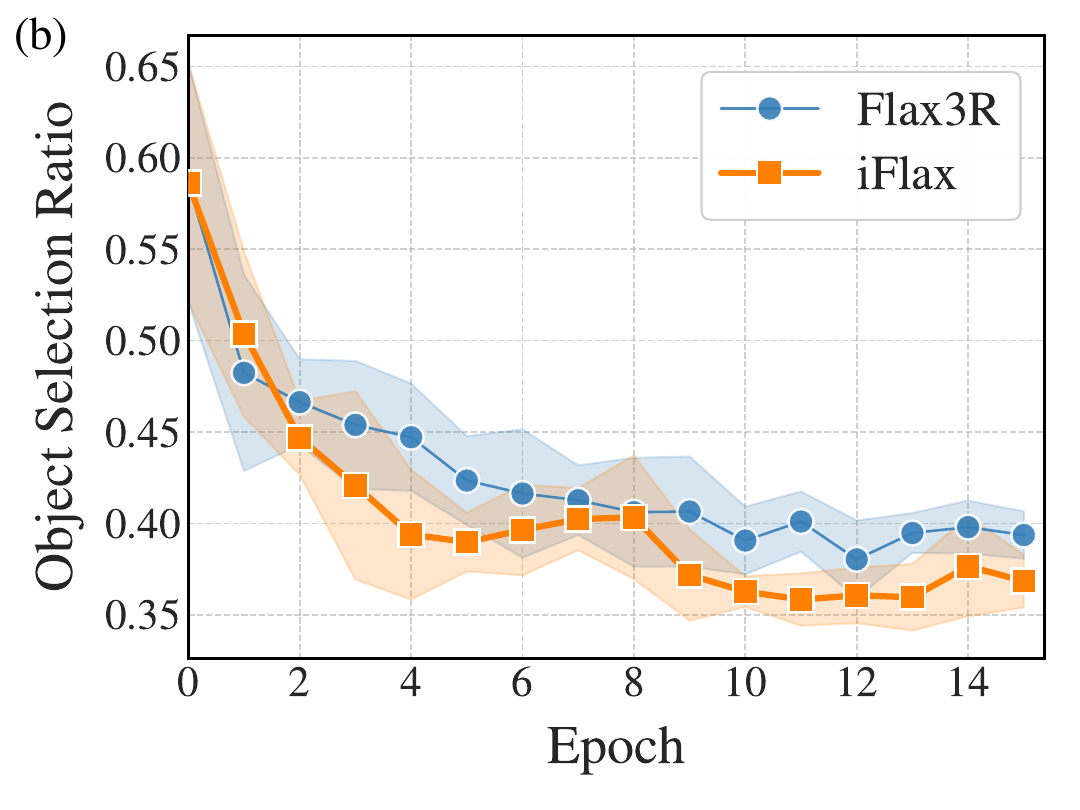}
        \end{overpic}%
        \label{fig:val_osr}%
    }
    \hfill
    \subfloat[{Search Efficiency}]{%
        \begin{overpic}[width=0.32\linewidth]{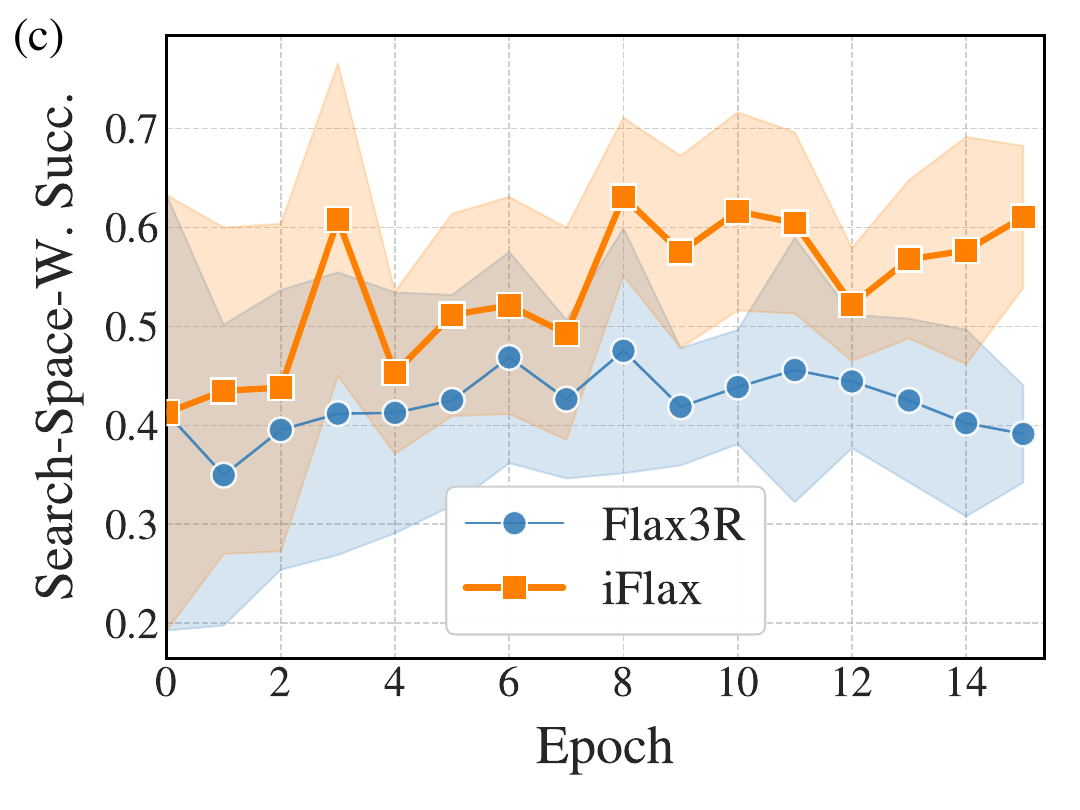}
        \end{overpic}%
        \label{fig:val_ssp}%
    }
    \caption{Validation trends on \textit{MazeNamo} over 15 epochs. \shortname is compared with the offline-trained Flax3R baseline in terms of (a) Solvability Within Limited Time, (b) Identification of Important Objects, and (c) Search Efficiency on the validation set. The shadows represent the variances over 10 random seeds.}
    \label{fig:validation_metrics}
\end{figure*}

\subsection{Ablation Studies}\label{subsec:ablation}
We study four planner variants in the \textit{MazeNamo} ablations:
\begin{itemize}
    \item Flax: the baseline method \cite{du2026fast}.
    \item \shortname: our full method.
    \item Flax3R: Flax augmented with our 3R recovery strategy.
    \item iFlax w/o 3R: \shortname without 3R recovery strategy.
\end{itemize}

\begin{table*}[!t]
    \caption{Ablation of bilevel optimization and 3R recovery on \textit{MazeNamo}.}
    \label{tab:maze_ablation}
    \centering
    \begin{tabular}{@{}C{0.08\linewidth}@{}C{0.1\linewidth}@{}*{8}{C{0.1\linewidth}@{}}}
        \toprule
        \multirow{2}{*}{Map Size} & \multirow{2}{*}{Level} & \multicolumn{2}{c}{Flax \cite{du2026fast}} & \multicolumn{2}{c}{\makecell[c]{Flax3R (Ours)}} & \multicolumn{2}{c}{\makecell[c]{iFlax w/o 3R (Ours)}} & \multicolumn{2}{c}{\makecell[c]{\shortname (Ours)}} \\
        & & \makecell[c]{FR $\downarrow$} & \makecell[c]{WPT (s) $\downarrow$} & \makecell[c]{FR $\downarrow$} & \makecell[c]{WPT (s) $\downarrow$} & \makecell[c]{FR $\downarrow$} & \makecell[c]{WPT (s) $\downarrow$} & \makecell[c]{FR $\downarrow$} & \makecell[c]{WPT (s) $\downarrow$} \\
        \midrule
        \multirow{4}{*}{10$\times$10} & Easy & 0.020 & 0.616 & 0.011 & 0.569 & 0.011 & 0.738 & \underline{\textbf{0.000}} & \underline{\textbf{0.470}} \\
        & Medium & 0.024 & 0.790 & 0.017 & 0.669 & 0.015 & 0.944 & \underline{\textbf{0.002}} & \underline{\textbf{0.596}} \\
        & Hard & 0.104 & 1.899 & 0.094 & 1.327 & 0.139 & 1.875 & \underline{\textbf{0.035}} & \underline{\textbf{1.040}} \\
        & Expert & 0.379 & 3.034 & 0.202 & 1.994 & 0.300 & 2.386 & \underline{\textbf{0.060}} & \underline{\textbf{1.124}} \\
        \midrule
        \multirow{4}{*}{12$\times$12} & Easy & 0.023 & 1.917 & 0.021 & 1.663 & 0.003 & 1.808 & \underline{\textbf{0.000}} & \underline{\textbf{1.199}} \\
        & Medium & 0.031 & 2.982 & 0.024 & 2.278 & 0.013 & 2.861 & \underline{\textbf{0.002}} & \underline{\textbf{1.658}} \\
        & Hard & 0.170 & 8.086 & 0.124 & 5.530 & 0.208 & 7.229 & \underline{\textbf{0.076}} & \underline{\textbf{3.734}} \\
        & Expert & 0.364 & 10.262 & 0.232 & 7.078 & 0.286 & 7.484 & \underline{\textbf{0.054}} & \underline{\textbf{3.101}} \\
        \midrule
        \multirow{4}{*}{15$\times$15} & Easy & 0.026 & 7.488 & 0.024 & 4.963 & 0.005 & 6.484 & \underline{\textbf{0.000}} & \underline{\textbf{3.602}} \\
        & Medium & 0.065 & 12.304 & 0.047 & 7.498 & 0.067 & 8.938 & \underline{\textbf{0.019}} & \underline{\textbf{4.829}} \\
        & Hard & 0.079 & 13.514 & 0.048 & 8.182 & 0.106 & 9.095 & \underline{\textbf{0.019}} & \underline{\textbf{4.702}} \\
        & Expert & 0.393 & 23.771 & 0.192 & 14.812 & 0.321 & 16.802 & \underline{\textbf{0.068}} & \underline{\textbf{7.765}} \\
        \midrule
        \multicolumn{2}{c}{\textit{Average}} & 0.140 & 32.14\% & 0.086 & 21.88\% & 0.123 & 26.59\% & \underline{\textbf{0.028}} & \underline{\textbf{13.78\%}} \\
        \bottomrule
    \end{tabular}
\end{table*}

\subsubsection{Overcoming Exposure Bias via Bilevel Optimization}
To test whether our online bilevel optimization overcomes the exposure bias of offline object-importance supervision, we first compare each offline-trained planner with its online-trained counterpart in Table~\ref{tab:maze_ablation} (Flax vs. iFlax w/o 3R; Flax3R vs. \shortname). In each pair, the key difference is whether the scorer is trained from fixed offline labels or from adaptive pseudo-supervision in the score-pruned search spaces. Bilevel optimization improves the 3R-equipped planner substantially, dropping the average failure rate from 8.6\% to 2.8\% and the normalized weighted planning time from 21.88\% to 13.78\%. This supports the claim that adaptive pseudo-supervision improves object scoring by closing the train-test mismatch of fixed offline labels.
To further verify this effect, we next analyze from three different aspects.

\paragraph{Solvability Within Limited Time}
To analyze how often the method fails to solve a validation task within the time budget, we track failure rate (FR) over training epochs.
The FR curve in \fref{fig:val_fr} shows that the gain from \shortname is consistent across training, not limited to a few isolated epochs. After starting from the same point as the offline baseline, the FR of \shortname drops faster in the first several epochs and then stays below Flax3R for the rest of training. The gap becomes clearer in the second half, where the FR of \shortname remains near a much lower plateau while the FR of Flax3R fluctuates at a higher level. This pattern matters because it suggests that adaptive pseudo-supervision does more than accelerating early learning. It keeps optimizing the scorer toward predicting object sets that the lower-level planner can solve more efficiently and reliably, which leads to fewer unsolved validation tasks.

\paragraph{Identification of Important Objects}
To further analyze how many objects \shortname keeps in the final object set when the planner succeeds, we introduce Object Selection Ratio (OSR), which is the ratio between the number of objects in the final simplified task and the number of objects in the original task, namely $N_{\text{final\_obj.}} / N_{\text{total\_obj.}}$. A lower OSR means the method solves the task with fewer objects. As shown in \fref{fig:val_osr}, both Flax3R and \shortname learn to select smaller active object sets as training proceeds, but \shortname lowers OSR faster and reaches a lower ratio, especially after the early epochs. 
This means \shortname learns to remove more unnecessary objects than the baseline. By jointly observing \fref{fig:val_fr} and \fref{fig:val_osr}, we can see that \shortname tends to select fewer objects while solving more tasks.

\paragraph{Search Efficiency}
To analyze the planning efficiency of \shortname more directly, we need a metric that reflects the search space itself and measures how close the explored search space is to the minimum required search space. 
We quantify search-space size by the number of state nodes the planner evaluates during search. For each task,  we need a reference value for the number of state nodes evaluated after removing unnecessary objects. To obtain this reference, we collect the objects involved in the final plan $\pi^\star$, reconstruct the task using only those necessary objects, and run the symbolic planner again on this reduced task.
Let $n^{\text{ori}}_{es}$ be the number of evaluated nodes in the original run, and let $n^{\text{nec}}_{es}$ be the number of evaluated nodes in the latter near-optimal run.
We therefore introduce Success Rate Weighted by Search Space (SSP):
\begin{equation}
\text{SSP}=\frac{1}{|\mathcal{D}_{\text{val}}|}\sum_{d\in\mathcal{D}_{\text{val}}}\mathbbm{1}[\text{success}_d]\min\!\left(\frac{n^{\text{nec}}_{es,d}}{n^{\text{ori}}_{es,d}}, 1\right),
\end{equation}
where $\mathbbm{1}[\text{success}_d]$ indicates whether task $d$ is solved, $\mathcal{D}_{\text{val}}$ is the set of solved tasks. 
A higher SSP means that the method not only succeeds more often, but also solves tasks with smaller search effort, i.e., a higher SSP means less wasted search. As shown in \fref{fig:val_ssp}, \shortname stays above Flax3R across all epochs, with a clear gap once training moves beyond the initial warm-up period. This shows that \shortname improves the efficiency of successful searches, not just the success count. Together, the trends of FR, OSR, and SSP curves in \fref{fig:val_fr}, \fref{fig:val_osr}, and  \fref{fig:val_ssp} share a coherent result: \shortname fails less, keeps fewer unnecessary objects, and needs less search effort.

\subsubsection{Training Stability from the 3R Recovery Strategy}
We next test whether the 3R recovery strategy improves training stability by stabilizing lower-level search and, in turn, providing reliable adaptive feedback for the scorer.
In the offline setting, the performances of Flax and Flax3R in \tref{tab:maze_ablation} show that adding 3R to Flax reduces the average failure rate from 14.0\% to 8.6\% and the normalized weighted planning time from 32.14\% to 21.88\%. This shows that a single repair route is not enough when iterative expansion either misses critical objects or becomes bloated with unimportant ones.
The same pattern appears under imperative learning. 
Comparing iFlax w/o 3R with \shortname in \tref{tab:maze_ablation}, we see that adaptive pseudo-supervision alone is not enough. Without 3R, the method remains brittle and often degrades on harder instances because unstable lower-level search produces unstable supervision. After adding 3R, the average failure rate drops further from 12.3\% to 2.8\%, and the normalized weighted planning time drops from 26.59\% to 13.78\%. This gap shows that 3R is not only a better recovery mechanism at test time, but also the key component that makes bilevel training stable enough to consistently improve the scorer across difficult instances.

\begin{table*}[thpb]
    \caption{Effect of different relaxation rules (RR) in \shortname on expert $15\!\times\!15$ \textit{MazeNamo} tasks.}
    \label{tab:rr_ablation}
    \centering
    \begin{tabular}{@{}C{0.13\linewidth}@{}*{12}{C{0.069\linewidth}@{}}}
        \toprule
        \multirow{2}{*}{Map Size \& Level} & \multicolumn{2}{c}{PLOI \cite{silver2021planning}} & \multicolumn{2}{c}{\shortname (RR1)} & \multicolumn{2}{c}{\shortname (RR2)} & \multicolumn{2}{c}{\shortname (RR3)} & \multicolumn{2}{c}{\shortname (RR4)} & \multicolumn{2}{c}{\shortname (RR5)} \\
        & \makecell[c]{FR $\downarrow$} & \makecell[c]{WPT (s) $\downarrow$} & \makecell[c]{FR $\downarrow$} & \makecell[c]{WPT (s) $\downarrow$} & \makecell[c]{FR $\downarrow$} & \makecell[c]{WPT (s) $\downarrow$} & \makecell[c]{FR $\downarrow$} & \makecell[c]{WPT (s) $\downarrow$} & \makecell[c]{FR $\downarrow$} & \makecell[c]{WPT (s) $\downarrow$} & \makecell[c]{FR $\downarrow$} & \makecell[c]{WPT (s) $\downarrow$}\\
        \midrule
        15$\times$15 (Expert) & 0.526 & 24.40 & \underline{\textbf{0.068}} & \underline{\textbf{7.77}}& 0.167 & 11.93 & 0.157 & 11.55 & 0.138 & 11.02 & 0.129 & 10.96 \\
        \bottomrule
    \end{tabular}
\end{table*}

\begin{table}[t]
    \caption{Contribution of each planning stage to the solved instances across the evaluation domains.}
    \label{tab:pipeline_contribution}
    \centering
    \begin{tabular}{@{}C{0.3\linewidth}@{}*{3}{C{0.2\linewidth}@{}}}
        \toprule
        Domain & \textit{MazeNamo} & \textit{Soko.Plus} & \textit{Logi.Plus} \\
        \midrule
        Total & 21000 & 2500 & 1000 \\
        \midrule
        Iterative Expansion & 17030 & 556 & 309 \\
        \midrule
        Parallel Recovery & & & \\
        \quad \textit{Repair} & 227 & 219 & 35 \\
        \quad \textit{Restart} & 2601 & 566 & 278 \\
        \quad \textit{Rollback} & 781 & 258 & 269 \\
        \midrule
        Unsolved & 361 & 901 & 109 \\
        \bottomrule
    \end{tabular}
\end{table}

\subsubsection{Effects of Score-Guided Iterative Expansion and 3R Strategy}
To illustrate the importance of the proposed score-guided iterative expansion and 3R strategy, we track the \emph{final successful stage} for each task and aggregate the counts over 10 random runs in \tref{tab:pipeline_contribution}. These categories are mutually exclusive. A task is counted under \textit{Iterative Expansion} only if the initial expansion phase solves it before any recovery branch is launched. If that initial phase gets stuck and the planner enters 3R recovery, the task is then counted under the branch that eventually solves it, namely \textit{Repair}, \textit{Restart}, or \textit{Rollback}. If no stage succeeds, the task is counted as \textit{Unsolved}.

Under this counting rule, the initial expansion resolves a large fraction of the tasks: 17,030 in \textit{MazeNamo}, 556 in \textit{SokoMindPlus}, and 309 in \textit{LogisticsPlus}. This shows that, when the scorer is accurate enough, the planner often succeeds before any recovery is needed. However, once this initial phase gets stuck, the 3R recovery stage becomes essential.

Among the three branches, \textit{Restart} contributes the most, resolving 2,601 expansion-stuck cases in \textit{MazeNamo} and 566 in \textit{SokoMindPlus}. This suggests that resetting the active object set is a reliable way to escape severe search-space bloat after score-guided expansion has drifted off course. \textit{Rollback} is the second most effective branch, with 781, 258, and 269 recoveries across the three domains, showing the value of conservative backtracking near the ``boundary'' of tractability. \textit{Repair} contributes fewer resolutions, but it remains useful when the failure is local and a broad reset would be unnecessary. These results show that score-guided expansion and 3R play complementary roles: expansion handles the easy and moderate cases directly, while recovery resolves the hard cases where inaccurate scores would otherwise block planning.

\begin{figure}[t]
  \centering
  \includegraphics[width=0.48\textwidth]{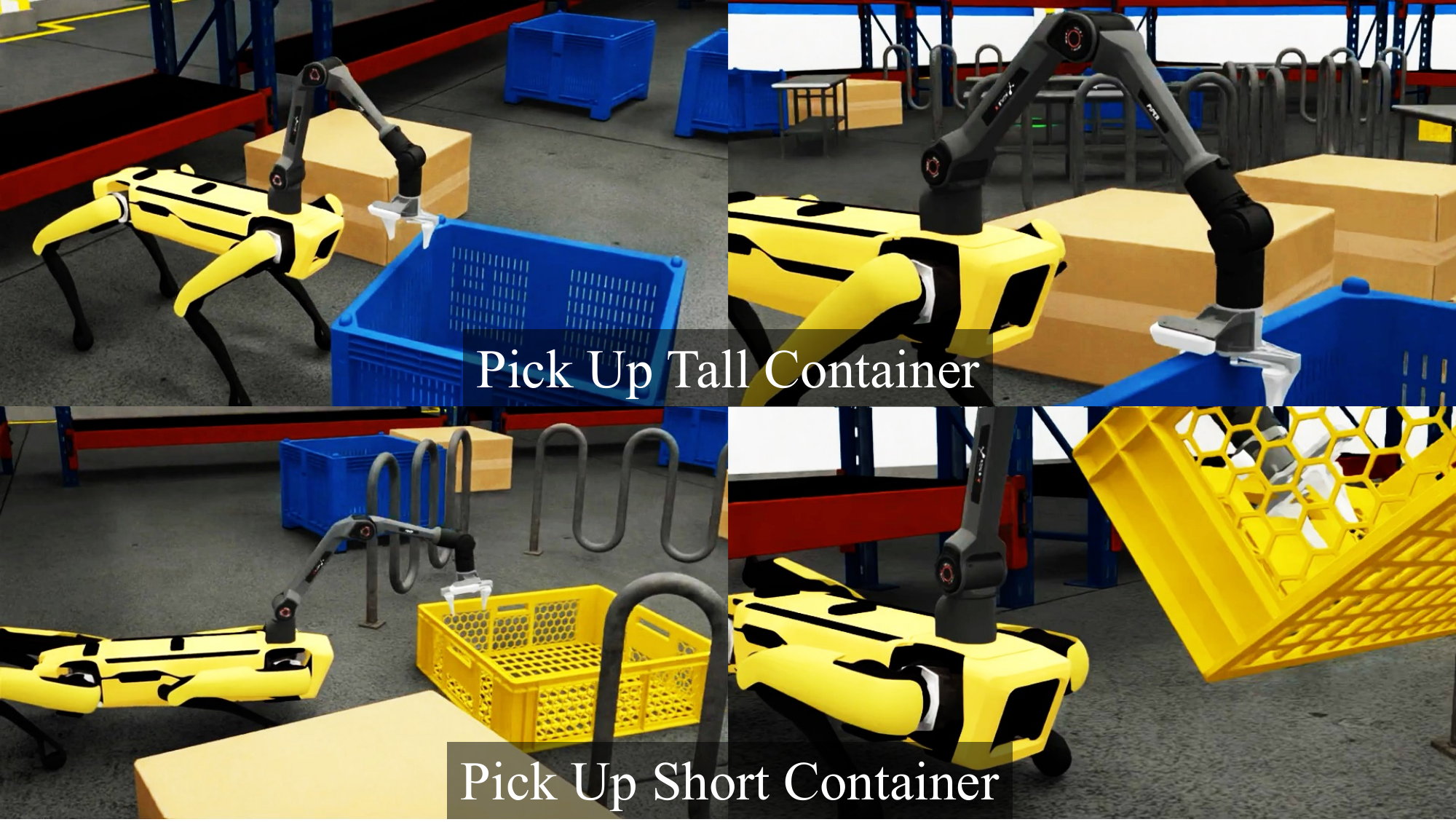}
  \caption{Isaac Sim pickup skills for tall and short containers. The robot needs to stand to pick up tall containers and sit to pick up short containers.
  % matching the symbolic manipulation constraints.
  }
  \label{fig:isaac_pickup}
\end{figure}

\subsubsection{Impact of Relaxation Rules}
To evaluate the role of domain-specific relaxation in the \textit{Repair} and \textit{Restart} stages, we test five relaxation strategies on expert $15\!\times\!15$ \textit{MazeNamo} tasks. These strategies remove all light obstacles (RR1), remove all movable obstacles (RR2), remove all heavy obstacles (RR3), treat all movable obstacles as light obstacles (RR4), or remove all obstacles (RR5). As shown in \tref{tab:rr_ablation}, rule quality matters: RR1 reaches a 6.8\% failure rate, whereas RR2 reaches 16.7\%. Even so, all relaxed variants outperform the non-relaxed baseline PLOI. This suggests that a useful relaxation does not need to be perfect. It only needs to simplify the task while preserving enough causal structure to guide recovery.

\begin{figure*}[thpb]
  \centering
  \includegraphics[width=\textwidth]{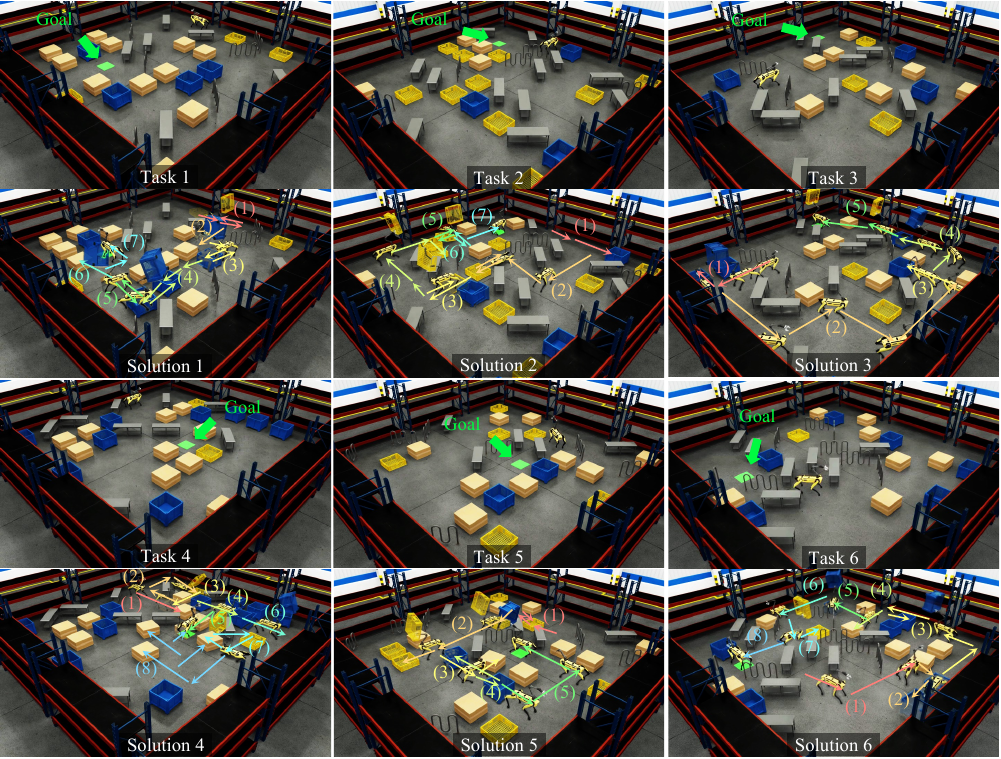}
  \caption{Representative Isaac Sim \textit{MazeNamo} tasks. Each pair shows the initial task and the corresponding execution trajectory.}
  \label{fig:isaac_tasks}
\end{figure*}

\subsection{Simulation Demonstrations}
We next address an important question for deployment: can the same planning model, trained on existing benchmarks, generalize to new logical constraints and support complete execution under domain differences?
Inspired by \textit{MazeNamo}, we generate challenging tasks in Isaac Sim \cite{NVIDIA_Isaac_Sim}, where movable obstacles include heavy boxes, tall containers, and short containers.
A key difference of this simulated challenge from \textit{MazeNamo} is that the symbolic domain encodes more logical constraints: (1) containers can only be placed on the ground, (2) the robot must stand to pick a tall container, and (3) it must sit to pick a short container, as shown in \fref{fig:isaac_pickup}. 

We convert each simulated scene and task into a PDDL problem, solve it with \shortname, and execute the returned high-level plan sequentially.
To isolate the planner-to-executor interface from perception uncertainty, we read the robot pose and each object's 3D bounding box directly from the simulator.
We then ground each high-level action with a predefined skill that includes a closed-loop controller and a behavior-tree-based recovery policy \cite{colledanchise2016behavior}.
The high-level goal in these tasks is ``\textit{robot navigation to a target location}.''

\paragraph*{Long-horizon Plan}
Figure~\ref{fig:isaac_tasks} shows representative problems, where ``(number)'' denotes the execution order and their colors denote the plan stages.
Note that a stage may contain multiple ordered high-level skills, and one skill may contain multiple ordered primitive steps.
For instance, the symbolic plan in Task 6 contains 57 ordered skills for interacting with three heavy boxes, four tall containers, and one short container.
A skill such as picking up a short container requires multiple steps, e.g., approach the container, sit down, detect it, grasp it, stand up, and retry if the grasp fails.
This leads to a very long-horizon task plan.
It is worth noting that \shortname is only trained on \textit{MazeNamo}, which is different from this domain because of the robot's kinematic constraints, e.g., the robot has to sit before picking up short containers due to limited arm length.
Even under these additional logical constraints, \shortname still solves 96\% of the tasks within the same time budget, which further demonstrates its generalizability.
The videos of these results can be found in the supplementary material.

\subsection{Real-World Demonstrations}
We then test whether the full perception-planning-execution pipeline supports complete hardware execution under sensing, geometry, and control constraints.
Inspired by the simulated study, we deploy \shortname on a real-world mobile manipulator in cluttered office and warehouse-style environments, where task-relevant objects include heavy boxes, short containers, and bottles.
A key difference of this real-world study from the Isaac Sim setting is that the system must first build the symbolic problem from sensed geometry and then execute the returned plan under real hardware uncertainty.

\subsubsection{Real-World Robot Platform}
\begin{figure*}[t]
  \centering
  \includegraphics[width=\textwidth]{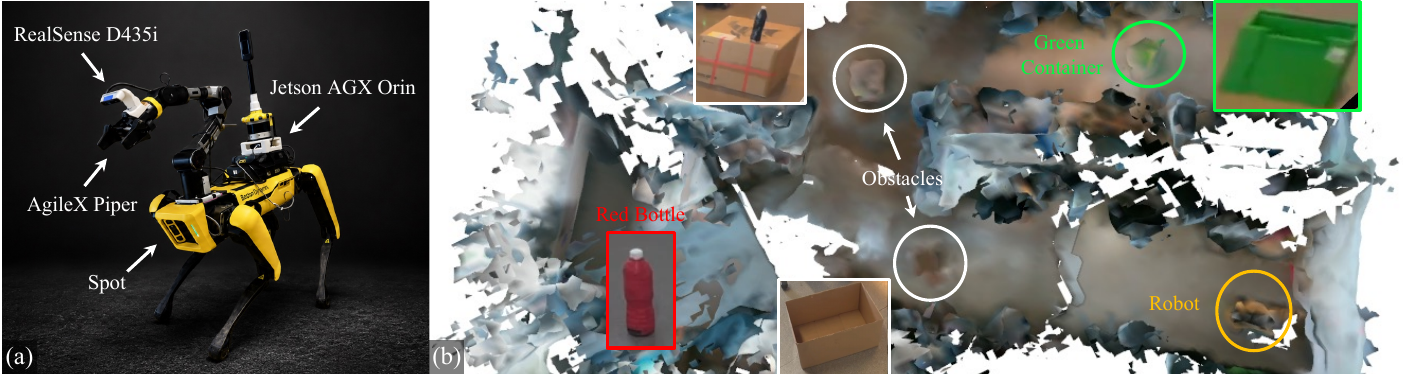}
  \caption{Real-world mobile manipulation platform and reconstructed office scene. (a) The deployed platform. (b) The registered office scene graph used for planning, and the insets highlight representative task-relevant objects in that scene.}
  \label{fig:real_world_platform}
\end{figure*}

In the real world, we deploy the system on a Spot quadruped with a computing board Jetson AGX Orin, an AgileX Piper manipulator, and a wrist-mounted RealSense D435i, as shown in \fref{fig:real_world_platform}a.
We first construct a consistent scene representation. Spot scans the empty environment to build a reference map for localization. After clutter is placed, we use a handheld platform with a Jetson AGX Thor and a RealSense D455 to scan the full scene, then reconstruct a scene graph with XRSLAM \cite{xrslam,li2024rd} and Hydra \cite{hughes2022hydra,hughes2024foundations}. 
An example of the simplified planning scene graph is shown in \fref{fig:real_world_platform}b, where we refine and register \cite{wang2023pypose} it to the map built by Spot so that detected obstacles and symbolic locations share a common coordinate frame.

\begin{figure*}[t]
  \centering
  \includegraphics[width=\textwidth]{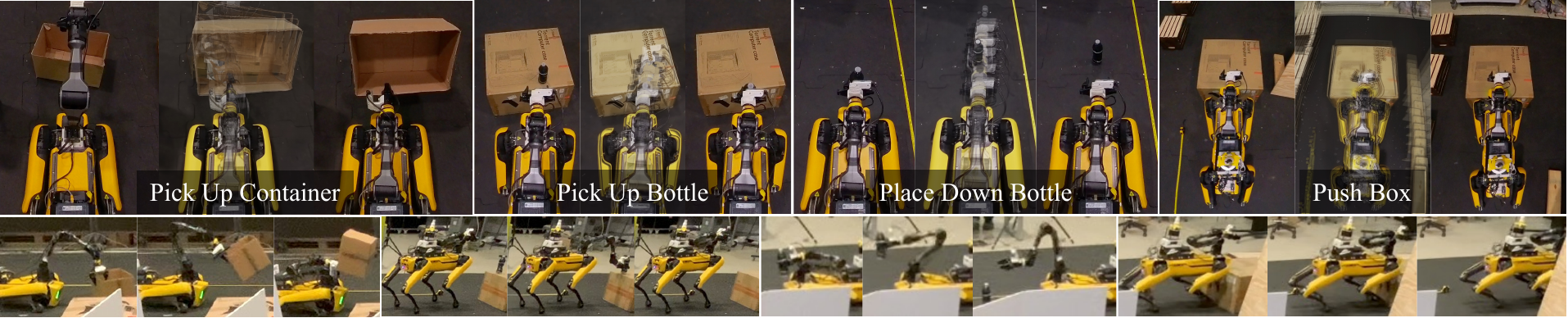}
  \caption{Representative real-world manipulation skills including container pickup, bottle pickup, bottle placement, and box pushing. Each skill consists of multiple ordered steps executed by a closed-loop controller with a behavior-tree-based recovery policy.}
  \label{fig:real_world_skills}
\end{figure*}

\begin{figure*}[t]
  \centering
  \includegraphics[width=\textwidth]{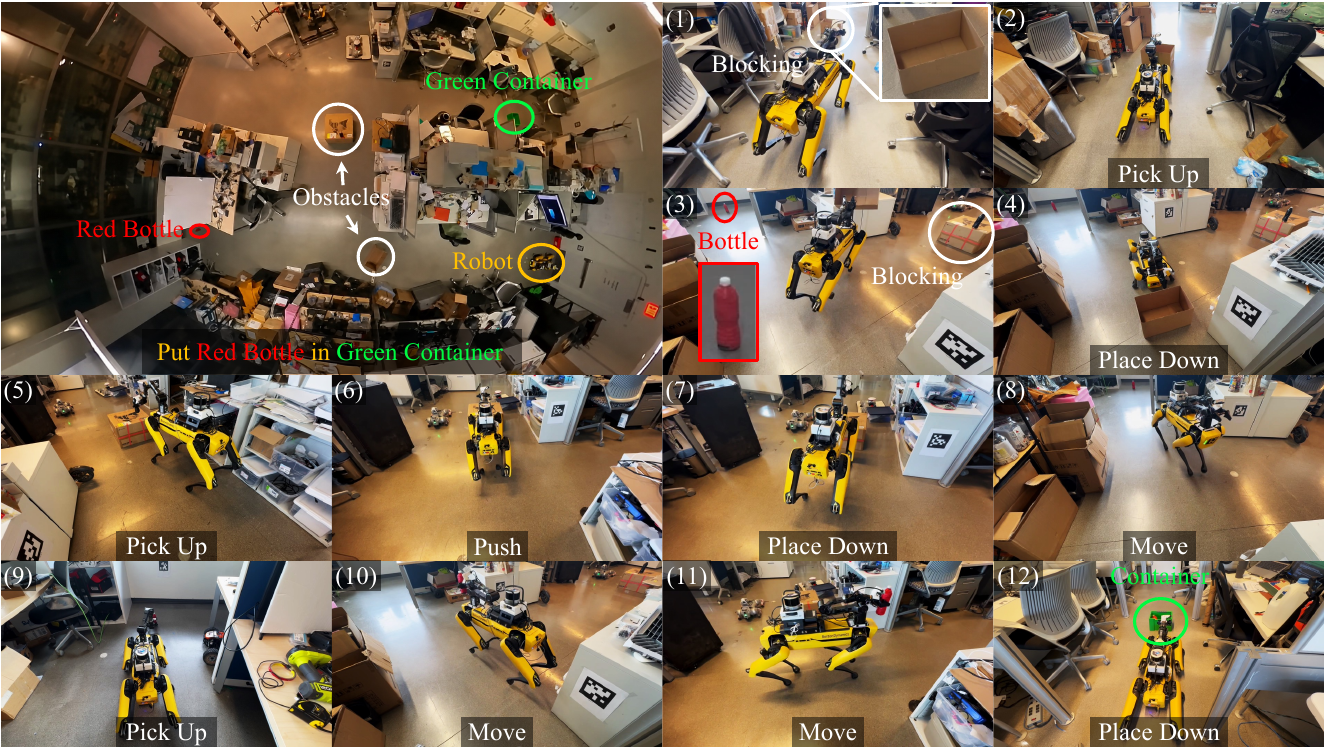}
  \caption{Real-world office task execution. The robot places the red bottle into the green container by removing obstacles (a pickable container, a pickable bottle, and a pushable box), moving items, and navigating between manipulation sites.}
  \label{fig:office_full_task}
\end{figure*}

Similar to the pipeline in Isaac Sim, we translate scene graphs into PDDL, solve the task with \shortname, and execute the plan with a library of grounded skills for interaction with heavy boxes, short containers, and bottles. 
Accordingly, the deployed skill set includes container and bottle pickup and placement, as well as box pushing, as shown in \fref{fig:real_world_skills}. 
The logical constraints mirror the physical task structure: bottles can be placed on the ground or on heavy boxes, containers can only be placed on the ground, and the robot must sit to pick a short container. 
Specifically, we use WildDet3D \cite{huang2026wilddet3d} to estimate 3D bounding boxes, a grid-search heuristic to propose grasp poses, and MoveIt~2 \cite{coleman2014reducing} with KDL-based inverse kinematics \cite{smits2011kdl} for collision-free manipulator planning.

\subsubsection{Real-World Mobile Manipulation in Office}
To test whether the real-world pipeline can solve a mobile manipulation goal in a daily cluttered scene, we show a representative office task in \fref{fig:office_full_task}. The goal is to ``\textit{place the red bottle into the green container}.'' Unlike a direct pick-and-place, this task couples object transport with access constraints: a pickable container, a pickable bottle, and a pushable box block the relevant paths and create a strict order for access and grasping.
The plan therefore contains several ordered navigation and manipulation skills. As shown in \fref{fig:office_full_task}, the robot must (1--4) first remove the blocking container, (5) pick up the bottle on the box, (6) push the box to clear the workspace, (7) place down the bottle on the box, (8) revisit the manipulation site, and then (9--12) transport the target bottle to the goal. These interactions with obstacles along the way matter because they create the free space and grasp access required later in the task. The successful execution shows that \shortname can reason over intervening obstacle management, not only the final object arrangement, in real-world long-horizon mobile manipulation.

\begin{figure*}[t]
  \centering
  \includegraphics[width=\textwidth]{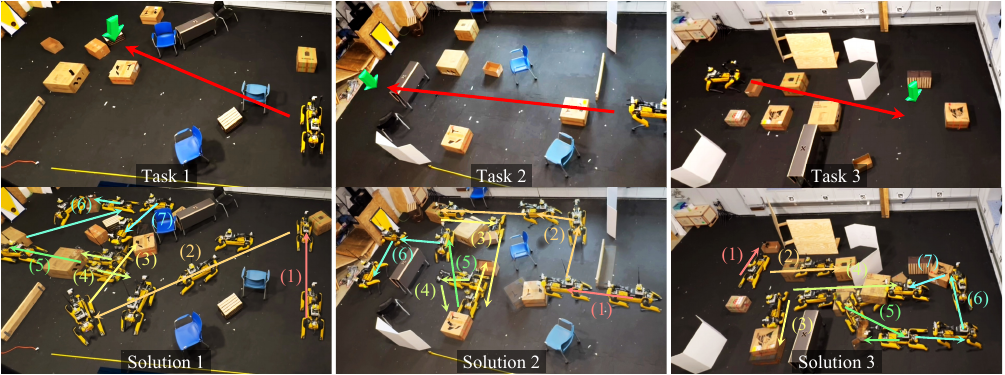}
  \caption{Representative real-world warehouse navigation tasks. The three tasks all use the goal ``\textit{reach location},'' with large green arrows marking the goals.}
  \label{fig:real_world_tasks_1}
\end{figure*}

\begin{figure*}[t]
  \centering
  \includegraphics[width=\textwidth]{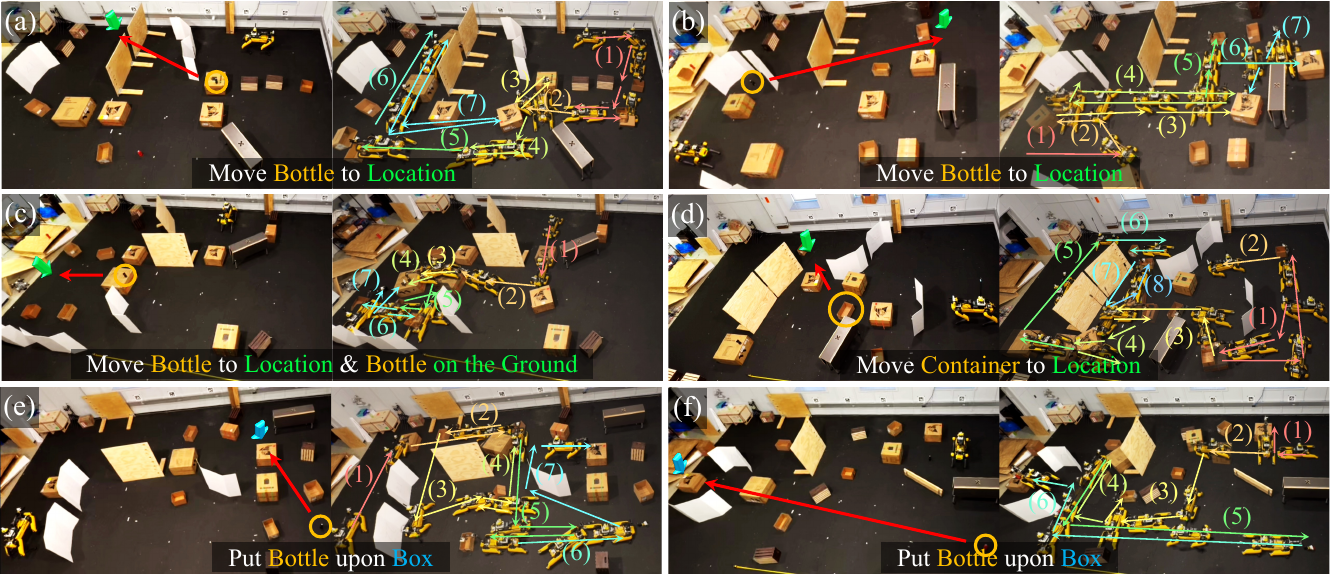}
  % \vspace{-20pt}
  \caption{Representative real-world warehouse mobile manipulation tasks. The six tasks cover ``\textit{move bottle to location},'' ``\textit{move bottle to location, bottle on the ground},'' ``\textit{move container to location},'' and ``\textit{put bottle upon box}.'' Large arrows mark task goals, and the overlaid robots illustrate the execution sequence.}
  \label{fig:real_world_tasks_2}
\end{figure*}

\subsubsection{Real-World Warehouse Navigation}
To test whether \shortname can solve real-world ``\textit{reach location}'' tasks in larger cluttered environments, we demonstrate its performance in warehouse navigation.
We show three examples in \fref{fig:real_world_tasks_1}, where all tasks share the same high-level goal, but each scene contains many movable objects, so the planner must identify which bottles, containers, and boxes actually determine access to the target location while ignoring unrelated clutter.
Across these navigation tasks, \shortname takes 0.92s on average to obtain a feasible plan.
Task 3 in \fref{fig:real_world_tasks_1} gives a representative example. The robot must manipulate two containers, push four boxes in a specific order, and move through the remaining clutter before it can reach the goal. This is not a single navigation segment, but a staged plan that alternates obstacle rearrangement and navigation. The successful executions show that \shortname can coordinate symbolic reasoning with grounded skills for long-horizon navigation among movable obstacles in the real world.

\subsubsection{Real-World Warehouse Mobile Manipulation}
To further test whether the same pipeline can solve mobile manipulation goals that specify \textit{where an item should be located} and \textit{what it should be placed on}, we demonstrate its performance in warehouse mobile manipulation. The goals include ``\textit{move bottle to location},'' ``\textit{move bottle to location and bottle on the ground},'' ``\textit{move container to location},'' and ``\textit{put bottle upon box}.'' These constraints make the task more difficult than ``\textit{reach location}'' by introducing requirements on \textit{where} and \textit{how} an item is placed.
For instance, ``\textit{move bottle to location and bottle on the ground}'' requires the bottle to reach the target location and remain on the ground at the end, while ``\textit{put bottle upon box}'' requires the bottle to be placed on a specific box.
Across these warehouse mobile manipulation tasks, \shortname takes 1.24s on average to obtain a feasible plan.

We show six tasks in \fref{fig:real_world_tasks_2}, where a representative example is Task (d): ``\textit{move container to location}.'' The plan contains 51 high-level skills. The robot (1) first moves Light Container 1 from the bottom right to the top right so that (2) it can push Heavy Box 2 downward and open the corridor to the target container. It then (3) carries Light Container 2 (the target container) through a narrow passage, turns around, and sits to place it temporarily because the gripper must be free for the remaining item rearrangement. The robot cannot directly push Heavy Box 3 that blocks the goal location because there is not enough free space behind it, so it detours along the bottom path, (4) moves Bottle 1, (5) pushes Heavy Box 2, (6) removes Light Container 3 on the top left, and (7) pushes Heavy Box 3 downward to connect Light Container 2 with the goal location. Finally, it (8) picks up Light Container 2 again and places it at the goal.
Among these bottles, containers, and boxes in the scene, the plan manipulates only the few objects that are necessary to get access to the target container and then clear a feasible route to the goal. These executions therefore highlight the central capability of \shortname: identifying the task-relevant objects and ordering the necessary obstacle interactions under complex logical constraints.
The successful deployments show that this planning capability remains effective beyond the original \textit{MazeNamo} benchmark and transfers to real-world long-horizon mobile manipulation. The videos of these results can be found in the supplementary material.
% Video results are available on the ....

\section{Conclusion and Limitations}
We presented \shortname, a neuro-symbolic framework for long-horizon task planning under complex logical constraints. The core idea is to learn object importance as a bilevel optimization problem, so the scorer is trained from planner feedback in the same pruned search spaces it creates at deployment. We also introduced parallel 3R recovery, namely \textit{Repair}, \textit{Restart}, and \textit{Rollback}, to keep lower-level planning stable when pruning errors occur.
Experiments on \textit{MazeNamo}, \textit{SokoMindPlus}, and \textit{LogisticsPlus} show consistent gains in both robustness and efficiency. On \textit{MazeNamo}, \shortname reduces failure rate by 80.04\% and weighted planning time by 57.14\% relative to Flax, while the ablations show that both imperative bilevel learning and 3R are necessary for the full improvement. We also validated the framework in Isaac Sim and on a quadruped-based mobile manipulator, showing end-to-end execution from symbolic problem construction to skill-level control.

Nevertheless, the current study has several limitations. First, like most PDDL-based planning systems, \shortname depends on a pre-defined symbolic domain, discrete predicates, and a finite action vocabulary. 
As a result, it cannot directly model continuous robot motion or dynamics constraints. In our real world experiments, task planning and motion planning therefore remain separate, and the task planner may not fully account for motion-level feasibility or execution cost when selecting symbolic actions.
Although this paper focuses on task planning, \shortname's self-supervised training and recovery strategies may offer useful insights for improving the efficiency and robustness of future integrated task and motion planning systems.
Second, this paper focuses on feasibility and planning efficiency rather than formal guarantees on planning optimality or resiliency under rapidly changing environments. 
This emphasis is motivated by our observation that, in real world deployments, efficiently finding a feasible plan can be more critical than guaranteeing an optimal planning solution. In future work, we will further investigate task planning within dynamic and uncertain environments, together with tighter integration with lower-level motion planning.

% \section*{Acknowledgments}
% This should be a simple paragraph before the References to thank those individuals and institutions who have supported your work on this article.

% {\appendix[Proof of the Zonklar Equations]
% Use $\backslash${\tt{appendix}} if you have a single appendix:
% Do not use $\backslash${\tt{section}} anymore after $\backslash${\tt{appendix}}, only $\backslash${\tt{section*}}.
% If you have multiple appendixes use $\backslash${\tt{appendices}} then use $\backslash${\tt{section}} to start each appendix.
% You must declare a $\backslash${\tt{section}} before using any $\backslash${\tt{subsection}} or using $\backslash${\tt{label}} ($\backslash${\tt{appendices}} by itself
%  starts a section numbered zero.)}


\bibliographystyle{IEEEtran}
% \bibliography{IEEEabrv,../bib/paper}
\bibliography{IEEEabrv,refs/references}

\ifdefined\PaperBios
  % Replace fig1.png and the placeholder biography text with each author's final photo and biography.
\begin{IEEEbiography}[{\includegraphics[width=1in,height=1.25in,clip,keepaspectratio]{fig1.png}}]{Qiwei Du}
Replace this placeholder with the final TRO biography for Qiwei Du.
\end{IEEEbiography}

\begin{IEEEbiography}[{\includegraphics[width=1in,height=1.25in,clip,keepaspectratio]{fig1.png}}]{Zitong Zhan}
Replace this placeholder with the final TRO biography for Zitong Zhan.
\end{IEEEbiography}

\begin{IEEEbiography}[{\includegraphics[width=1in,height=1.25in,clip,keepaspectratio]{fig1.png}}]{Shaoshu Su}
Replace this placeholder with the final TRO biography for Shaoshu Su.
\end{IEEEbiography}

\begin{IEEEbiography}[{\includegraphics[width=1in,height=1.25in,clip,keepaspectratio]{fig1.png}}]{Bowen Li}
Replace this placeholder with the final TRO biography for Bowen Li.
\end{IEEEbiography}

\begin{IEEEbiography}[{\includegraphics[width=1in,height=1.25in,clip,keepaspectratio]{fig1.png}}]{Yi Du}
Replace this placeholder with the final TRO biography for Yi Du.
\end{IEEEbiography}

\begin{IEEEbiography}[{\includegraphics[width=1in,height=1.25in,clip,keepaspectratio]{fig1.png}}]{Zhipeng Zhao}
Replace this placeholder with the final TRO biography for Zhipeng Zhao.
\end{IEEEbiography}

\begin{IEEEbiography}[{\includegraphics[width=1in,height=1.25in,clip,keepaspectratio]{fig1.png}}]{Taimeng Fu}
Replace this placeholder with the final TRO biography for Taimeng Fu.
\end{IEEEbiography}

\begin{IEEEbiography}[{\includegraphics[width=1in,height=1.25in,clip,keepaspectratio]{fig1.png}}]{Sebastian Scherer}
Replace this placeholder with the final TRO biography for Sebastian Scherer.
\end{IEEEbiography}

\begin{IEEEbiography}[{\includegraphics[width=1in,height=1.25in,clip,keepaspectratio]{fig1.png}}]{Jiaoyang Li}
Replace this placeholder with the final TRO biography for Jiaoyang Li.
\end{IEEEbiography}

\begin{IEEEbiography}[{\includegraphics[width=1in,height=1.25in,clip,keepaspectratio]{fig1.png}}]{Chen Wang}
Replace this placeholder with the final TRO biography for Chen Wang.
\end{IEEEbiography}

\fi

\vfill

\end{document}